\documentclass[acmsmall, authordraft=false]{acmart}

\usepackage{changepage}
\usepackage{subcaption}
\usepackage{pdflscape}
\usepackage{xcolor}
\definecolor{footnotered}{RGB}{255,0,0}
\makeatletter
\renewcommand\@makefnmark{{\color{footnotered}\textsuperscript{\normalfont\@thefnmark}}}
\makeatother

\acmConference[XY'99]{XX Conference}{April
11--20}{Singapore}
\copyrightyear{2025}
\setcopyright{none}
\renewcommand\footnotetextcopyrightpermission[1]{}

\AtBeginDocument{%
  }

\begin{document}

\title{Tastes without distinction: silicon samples and the synthetic construction of tastes.}

\author{Xiangyu MA}
\email{xy.ma@ntu.edu.sg}
\affiliation{%
  \institution{Nanyang Technological University}
  \country{Singapore}
}

\author{Mengmi ZHANG}
\email{mengmi.zhang@ntu.edu.sg}
\affiliation{%
  \institution{Nanyang Technological University}
  \country{Singapore}
}

\author{Shannon ANG}
\email{shannon.ang@ntu.edu.sg}
\affiliation{%
  \institution{Nanyang Technological University}
  \country{Singapore}
}

\author{Minne CHEN}
\email{minne.chen@ntu.edu.sg}
\affiliation{%
  \institution{Nanyang Technological University}
  \country{Singapore}
}

\begin{abstract}
  \noindent Large-language models have proven to be remarkable if inconsistent parrots of public attitudes and opinions.
The extent to which LLMs are able to produce reasonable approximations of cultural taste remains an open empirical question that becomes more urgent by the day, with market research companies already offering provisional `synthetic' survey panels and the contamination of standard survey data from LLM-generated responses.
In this study, we build on past work on silicon sampling by extending considerations of their ecological, relational, and positional fidelity in the doomain of cultural tastes.
We use large-language models from OpenAI, Anthropic, and DeepSeek to produce 554,940 silicon surrogates of survey respondents from the \emph{Survey of Public Participation in the Arts} (SPPA).
We find these silicon surrogates' tastes to be highly stylized facsimiles of human tastes.
First, silicon samples are super-omnivorous with a systematic postive-bias for \emph{liking}.
These individual-level bias of silicon samples are not well-explained by the WEIRD-bias often discussed in the literature.
Second, the complex relationality in real taste structures is completely distorted among silicon samples.
Third, very little of the known cultural alignment between tastes and social space are preserved.
Silicon samples juvenilize age-taste associations, resurrect anachronistic class-taste associations, and caricaturize gender- and race-taste associations.

\vspace{\baselineskip}


\end{abstract}

\maketitle
\pagestyle{plain}

\pagebreak
\hypertarget{introduction}{%
\section{Introduction}\label{introduction}}

\begin{flushright}
  \begin{minipage}{0.5\linewidth}
    \itshape\raggedright
    Let me say and not mourn: //
    the world lives in the death of speech //
    and sings there.
    \par\medskip
    \upshape\raggedleft
    --- Wendell Berry, \emph{Silence}
    \par
  \end{minipage}
\end{flushright}
\bigskip

\noindent Large-language models have proven to be remarkable if inconsistent parrots of public attitudes and opinions (e.g \citealt{argyle_out_2023,bisbee_synthetic_2024,kozlowski_silico_2024}).
The extent to which LLMs are able to produce accurate surrogates of cultural taste and consumption remains an open empirical question that becomes more urgent by the day, with market research companies already offering provisional ``synthetic'' survey panels (e.g. \citealt{currie_qualtrics_2025,verasight_can_2026}).
In this study, we build on past work on silicon sampling by extending considerations of its algorithmic fidelity and alignment to the domain of cultural tastes.
We use large-language models from OpenAI, Anthropic, and DeepSeek to produce silicon surrogates of respondents from the \emph{Survey of Public Participation in the Arts} (SPPA), compare the tastes and preferences from these silicon samples against their SPPA counterparts.
We evaluate the ecological and relationality fidelity of these silicon tastes.
Then, using a mixed-effects meta-analysis framework, we evaluate the multivariate associations between silicon tastes and key dimensions of social space.

We find the following.
First, silicon samples are super-omnivorous with a systematic postive-bias for \emph{liking}.
These individual-level bias of silicon samples are not well-explained by the WEIRD-bias often discussed in the literature.
Second, the complex relationality in real taste structures is completely distorted among silicon samples.
Third, very little of the known cultural alignment between tastes and social space are preserved.
Silicon samples juvenilize age-taste associations, resurrect anachronistic class-taste associations, and caricaturize gender- and race-taste associations.
Silicon samples attenuate age-taste associations, resurrect anachronistic class-taste associations, caricaturize gender- and race-taste associations.

Assessing the fidelity of such silicon tastes is crucial for two main reasons.
First, if synthetic surrogates are able to -- even somewhat -- faithfully reproduce the eide of human groups, this dramatically expands the research design possibilities for quantitative studies of culture.
LLMs answer surveys quickly, cheaply, and without fatigue or complaint.
Synthetic panels make it possible to field surveys of cultural preference and consumption that were thought to be too long, laborious, or fine-grained.
Conventional surveys of cultural taste tend to rely on responses to fuzzy cultural categories, such as genres (e.g. \citealt{peterson_changing_1996, ma_what_2020}).
As Vandebroeck (\citealt{vandebroeck_thinking_2022}) points out, this convention assumes that consumers possess a taxonomy of cultural categories similar to the survey, and furter that the cultural categories themselves are not ambiguous subjects of contestation.
However, surveys that probe specific aspects and items constitutive of cultural taste are difficult to construct and tedious to complete.
It's easy to ask if a respondent likes trap or UK drill; much harder to persuade them to survey a battery of genre-specific artists.
Were silicon tastes to prove faithful, they would let researchers interrogate taste at the level of specific works, performances, or sub-genres, and to vary the framing of cultural categories experimentally, at a scale and cost that fielding such instruments on human respondents could never match.

Second, silicon sampling of consumer preferences and consumption is proliferating, notwithstanding scholarly reservations, making assessments of their fidelity an urgent priority.
While the social scientific community has urged caution when it comes to use of silicon samples (e.g. \citealt{gao_take_2025,bisbee_synthetic_2024,broska_mixed_2025}), these norms are unsettled and frequently transgressed.
It's important to have a better understanding of the quality, biases and limitations of silicon samples across a breadth of sociologically relevant domains (\citealt{lyman_balancing_2025}).
Researchers have used LLMs to simulate responses to contentious political items drawn from the American National Election Studies (\citealt{li_chatgpt_2024}), German public opinions (\citealt{ma_algorithmic_2025}), public opinions about climate change (\citealt{lee_can_2024}) to project audience attitudes and policy preferences (Wihbey and D'Alonzo 2025), and to model public-health attitudes such as vaccine acceptance (Khaokaew et al. 2025).
Importantly, a robust commercial interest in the provision of silicon samples has emerged.
Survey and market research companies like Qualtrics and VeraSight have already begun experimental rollouts of ``synthetic panels'' that are built using silicon samples (\citealt{isaza_can_2024,currie_qualtrics_2025, verasight_can_2026}).
Ordinary survey respondents also frequently turn to LLMs and chatbots to generate ostensibly `human' response surveys, meaning that even researchers who use orthodox survey methods are likely to face non-negligible contamination of `synthetic' responses in their data (\citealt{zhang_generative_2025,veselovsky_prevalence_2025}).
The faster silicon sampling diffuses across these settings, the more consequential its hidden biases become, and the more urgent it is to establish where it is, and is not, faithful.

\hypertarget{literature-review}{%
\section{Literature review}\label{literature-review}}

\subsection{Silicon sampling and the study of taste}
Large language models (LLMs) are probabilistic machine learning models that are used to process and produce textual data.
At a high level, LLMs resemble the autocomplete technologies that have become on search engines and other digital spaces, but with considerably greater scale and sophistication (\citealt{bail_can_2024}).
In this study, we extend ongoing work that applies LLMs to survey research.
In particular, we focus on the use of LLMs to simulate synthetic human samples in public opinion surveys, a practice often referred to as \emph{silicon sampling}.

The intuition behind silicon sampling is simple.
Homo silicus has advanced to such a state that many LLMs are able to converse and reason like humans, that the prospect of AI passing Turing tests just about seems a fait accompli (\citealt{dillion_can_2023, alvarado_turings_2026}).
If homo silicus really was capable of providing such a great facsimile of human responses, then it may be possible to productively sample responses from LLMs, the same way we would from human populations.
LLMs by the nature of their training and design are computational models of humans and likely possess a great deal of latent social information (\citealt{horton_large_2023}).
This latent information can often evince itself negatively in different guises of algorithmic bias, Argyle et al. (\citeyear{argyle_out_2023}) argue that it may be possible to instead think of them in productive terms as a model of the ``complex reflection of the many various patterns of association between ideas, attitudes, and contexts present among humans.''
The algorithmic fidelity of LLMs, when appropriately conditioned, means that researchers may be able to induce the model to produce outputs that correlate with the attitudes, opinions, and experiences of distinct human subpopulations.
Work is also underway exploring the applications of synthetic surrogates of human respondents in domains beyond survey research, from agent-based modeling, to psychological experiments, to behavioral economics experiments, to market research (\citealt{chen_emergence_2023,dillion_can_2023,horton_large_2023, park_generative_2023, goli_frontiers_2024}).

In our study, we advance ongoing work on the robustness and bounds of silicon sampling by applying to cultural fields, in particular exploring the application of silicon sampling to the case of public opinion surveys of cultural taste.
Cultural taste is commonly operationalized as a person’s first-order preferences, feelings, or desires towards an object in a cultural field (\citealt{ma_tastes_2024}).
We use the term ``silicon tastes'' to refer to the tastes and preferences expressed by silicon surrogates of human respondents.

\hypertarget{assessing-fidelity-silicon-tastes}{%
\subsection{Assessing the fidelity of silicon tastes}\label{assessing-fidelity-silicon-tastes}}

How faithful are silicon tastes to their human counterparts?
We contribute to ongoing work examing the bounds and robustness of silicon sampling by examing the quality and fidelity of synthetic cultural tastes.
We assess (1) the ecological fidelity of silicon tastes, and whether such validity is mediated by the so-called WEIRD bias of LLMs, (2) the higher-order relations among silicon tastes, and (3) the fidelity of taste-to-social-space associations inferred from silicon samples.
We ask three main research questions.
\begin{enumerate}
  \item How faithful are the ecological estimates of cultural tastes produced by silicon samples? And to what extent are discrepancies in estimates explained by the WEIRD-bias?
  \item To what extent do silicon samples preserve the relationality among tastes?
  \item To what extent are known cultural associations between tastes and social space preserved among silicon samples?
\end{enumerate}

To answer the first question, we compare the ecological estimates of cultural tastes for 17 genres of music from real and silicon samples.
Then, we compute and model the Jaccard distance between the real and silicon tastes.
To answer the second, we examine and compare the pairwise association matrices of real and silicon tastes.
To answer the third, we model the coefficient bias of silicon samples through a mixed-effects meta-regression.

\subsubsection{On the ecological fidelity of silicon tastes.}
Early work were quick to establish the surprising accuracy of silicon samples, especially when it came to ecological estimates.
In their seminal paper, \citealt{argyle_out_2023} first observed ``a high degree of correspondence between reported two-party presidential vote choice proportions from GPT-3 and ANES respondents.''
For instance, GPT-3 silicon respondents reported an aggregate 0.391 probability of voting for Mitt Romney for the 2012 US Presidential elections; the same percentage from the ANES was 0.404 (\citealt{argyle_out_2023}).
Bisbee et al. similarly created synthetic surrogates to the ANES, and find ``every synthetic mean falls within one standard deviation of the ANES average'' (\citealt[pg. 6]{bisbee_synthetic_2024}).
Studies of silicon samples in domains away from public opinion of politics have also found success.
Brand et al. use a silicon sampling approach to replicate conjoint studies of consumer preference, reporting that ``[o]verall, the resulting estimates are realistic, both in magnitudes and in distribution … a conjoint-like approach to preference estimation yields results that are strikingly similar to those found in a recent survey of real consumers conducted by Fong et al., as well as to additional human surveys we conduct'' (\citealt{brand_using_2023}).
Scholars have found similar success in contexts varying from environmental attitudes to moral judgements (\citealt{lee_can_2024,dillion_can_2023}).
The extent to which the same might apply to silicon panels of cultural tastes remains an open empirical question.

Further, we ask if discrepancies in ecological fidelity of silicon tastes may be attributable to what we term WEIRD-alignment bias (e.g. \citealt{bulte_llms_2025, liu_evaluating_2024}).
There can be considerable between-group variance in algorithmic fidelity, with divergences from real data particularly high among groups from communities that are underrepresented in the training data sets of today's foundation models (\citealt{madden_evaluating_2025,masoud_cultural_2025}).
The unevenness of the training data may result in the systematic misalignment of cultural tastes across the board, such that they are much more faithful for Wealthy, Educated, Industrialized, Rich and Democratic (WEIRD) sub-populations than for others (\citealt{xiao_algorithmic_2025, santurkar_whose_2023}).
We ask if such WEIRD-alignment bias might also explain systematic divergences in the ecological fidelity of silicon tastes.

\subsubsection{The relational fidelity of silicon tastes.}
The fidelity of silicon samples strike many as intuitively implausible; on closer examination, many researchers have uncovered many discrepancies between real survey samples and their silicon counterparts.
A persistent finding across studies is that the fidelity of silicon samples are restricted to the coarsest of ecological estimates (e.g. \citealt{bisbee_synthetic_2024}).
Dig any deeper, and you see clearly the discrepancies and human data.
For instance, researchers have noticed that silicon samples are uncannily precise in their estimates, underestimating the natural entropy to human responses (\citealt{bisbee_synthetic_2024,li_chatgpt_2024}).
In our case, we are primarily concerned with what we term \emph{relationality fidelity}.
Relational fidelity refers to the extent to which a silicon sample is able to faithfully reproduce extant relational associations among the variables, attributes and qualities in the reference data.
Relational fidelity is of particular importance to silicon tastes because tastes are fundamentally relational (\citealt{bourdieu_distinction:_1984,martin_what_2003, emirbayer_manifesto_1997}).
The social meaning of cultural objects, preferences, judgements, and tastes are developed in reference to one another; their relational properties lend them inherent meaning (\citealt{puetz_taste_2021, atkinson_class_2026}).
For example, the label of ``middle-brow'' that gets applied to certain cultural genres or objects is a relational label that requires the precedent construction of a high/low status bifurcation that is shared among a community, not an objective quality of said genre or object (\citealt{lizardo_reconceptualizing_2012,seabrook_nobrow:_2000}).
For a silicon surrogate to faithfully reproduce a reference individual's taste, it is not sufficient for them to coincide on one preference, but rather coincide on a broader set of preferences and versions (\citealt{goldberg_mapping_2011}).
For example, if taste $i$ (say, classic rock) goes with taste $j$ (say, blues) in the human data for a person $X$, then such covariations must also be observed in their silicon surrogate $\tilde X$.
The extent to which silicon samples preserve the relationality of tastes remains an open empirical question.

\subsubsection{The positional fidelity of silicon tastes.}
Positional fidelity refers to the extent to which a silicon samples faithfully preserves known associations between cultural tastes and positions in social space, such as class-, gender-, or race-based social positions.
There are other higher-order attributes of silicon samples that often fall-short.
Silicon samples often contain fundamentally different multivariate associations from real survey samples.
Compare, for instance, the regression coefficients inferred from silicon samples.
Bisbee et al. note that ``[w]hen we regress feeling thermometer scores on respondents' demographic attributes -- the type of analysis common in public opinion research -- the synthetic sample would frequently lead us to draw different inferences than if we relied on human respondents'' (\citealt[pg. 2]{bisbee_synthetic_2024}).
If this persists for cultural tastes, then, this may prove to be an enormous problem for usefulness of silicon samples in this domain.
Taste is known to be socially located.
In \emph{Distinction}, Bourdieu (\citeyear{bourdieu_distinction:_1984}) famously uses the case of taste to illustrate the dialectical relationship between subjective agency and social structure.
For this reason, tight taste-class homologies have been observed in a range of sociocultural contexts (e.g. \citealt{simmel_fashion_1957,accominotti_how_2018,atkinson_structure_2016}).
Such homologies between taste and social space are not limited to class alone, but also extend to other important organizing principles of social life, such as gender (\citealt{bourdieu_masculine_2001}), race (\citealt{atkinson_class_2026}), or life-stage (\citealt{ma_what_2020}).
We evaluate the extention to which such taste-space homologies are preserved among silicon samples.

\pagebreak
\hypertarget{data-and-methods}{%
\section{Data and methods}\label{data-and-methods}}

\begin{figure}
  \centering
  \caption{Overview of research design}\label{fig:pipeline}
  \includegraphics[width=\linewidth]{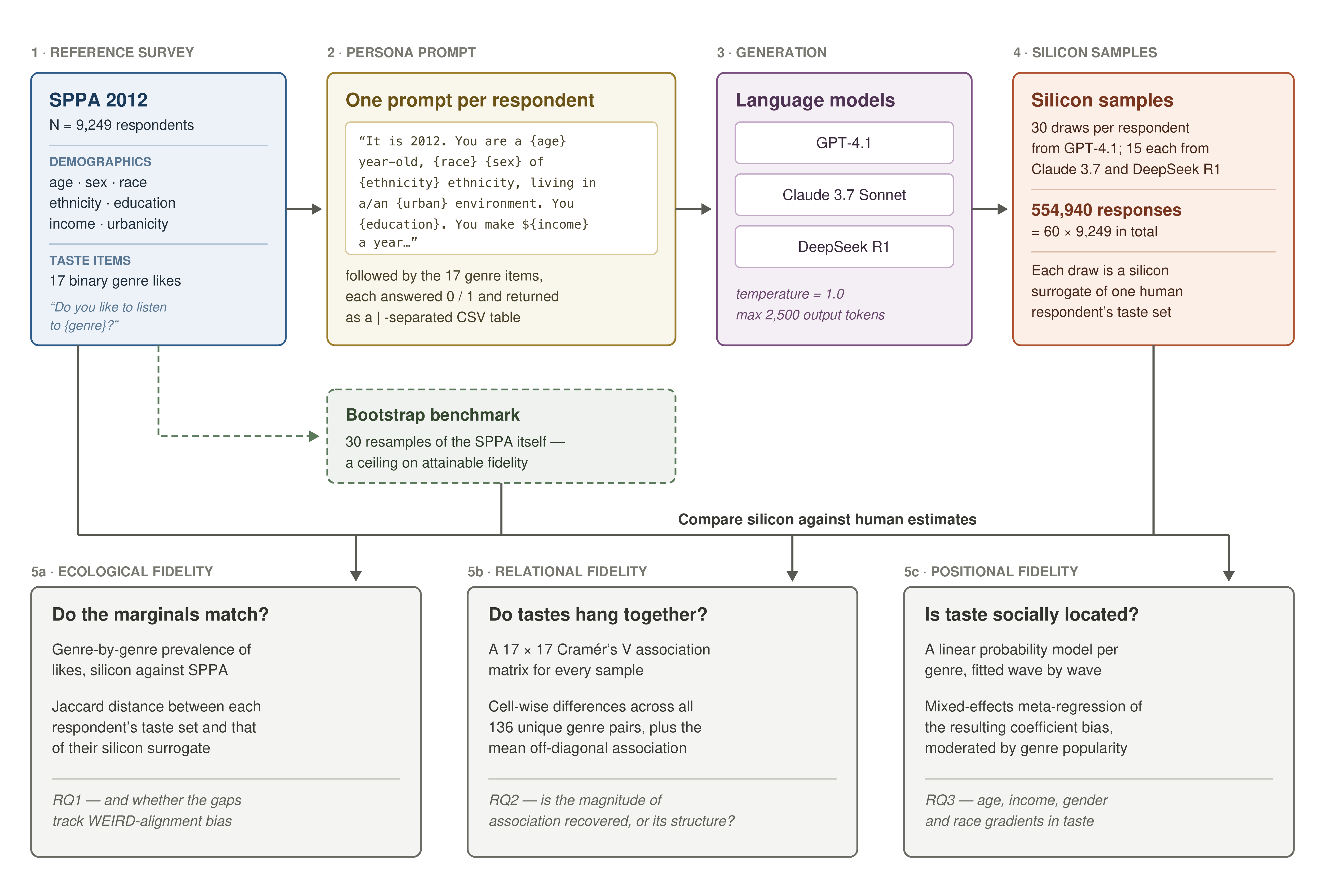}
\end{figure}

Figure~\ref{fig:pipeline} provides an overview of our research design.
We take the 2012 SPPA as our reference survey, convert each respondent's demographic profile into a persona prompt, and put that prompt to three large language models, drawing repeatedly from each to obtain a set of silicon surrogates for every human respondent.
We then compare the silicon and human samples along the three dimensions of fidelity set out above using bootstrap resamples of the SPPA itself as a benchmark for the degree of agreement attainable within the human data.

\hypertarget{data-source}{%
\subsection{Data source}\label{data-source}}

The SPPA is a repeated cross-sectional survey that measures cultural attitudes and participation in the United States.
From 1982 to 2022, the SPPA included items that tapped individuals' preferences for music genres, assessing respondents' likes for a range of musical genres.
Due to its longevity and representativeness, the SPPA serves as an effectively ``canonical'' data that that has provided the empirical basis for an enormous body of research into cultural tastes (e.g. \citealt{peterson_changing_1996}).

\hypertarget{generating-silicon-samples}{%
\subsection{Generating silicon samples}\label{generating-silicon-samples}}

We follow conventional approaches to silicon sampling to begin (e.g. \citealt{argyle_out_2023,bisbee_synthetic_2024}).
For a given survey $S$, we use an LLM model $L$ to generate a data set of matching synthetic responses $\hat S$.
After instructing the LLM model $L$ to adopt a persona defined by a set of respondent characteristics, we then ask it to answer the same set of prompts contained in $S$.\footnote{ Our choice of these `auxiliary' characteristics are constrained to those already present in the original SPPA questionnaire. Each persona is built from seven characteristics recorded for that respondent in the SPPA: age in years; sex, rendered as ``man'' or ``woman''; race, rendered as White, Black, Native, Asian, or mixed race; Hispanic ethnicity; educational attainment, rendered verbally across seven levels running from ``did not finish high school'' to ``have a doctorate degree''; annual household income, assigned the midpoint of the respondent's bracketed income category; and metropolitan residence, rendered as ``urban'' or ``rural.''}
We choose respondent personas and cultural taste questions to match questions that human respondents were asked in $S$, allowing us to gauge how accurately and precisely the synthetic LLM responses capture actual public opinion.
We generate synthetic samples using three different large-language models: (1) GPT-4.1 (\texttt{gpt-4.1-2025-04-14}), (2) Claude 3.7 Sonnet (\texttt{claude-3-7-sonnet-20250219}), and (3) DeepSeek R1 (\texttt{deepseek-reasoner}). For each of the models, we set temperature to be 1.0, and constrain the maximum output tokens to 2,500.

\hypertarget{prompt-settings}{%
\subsubsection{Prompt settings}\label{prompt-settings}}

We closely follow a style of prompting that has become dominant in political science, in particular when it comes to the simulation of synthetic attitudes towards political affairs and elections (e.g. \citealt{argyle_out_2023,bisbee_synthetic_2024}).
In short, we provide an LLM-powered conversational agent with a context, and then ask the agent to provide a response to a close-ended survey questionnaire, and to enclose the response in a \texttt{.csv} format.

Because LLM responses converge relatively quickly, we follow conventions and generate multiple synthetic responses for each real human respondent in the true survey (\citealt{bisbee_synthetic_2024}): 30 draws per respondent from GPT-4.1, and 15 draws per respondent from each of Claude 3.7 Sonnet and DeepSeek R1.
This yields a final data set of 554,940 ($60 \times 9{,}249$) synthetic responses.

\hypertarget{key-measures}{%
\subsection{Key measures}\label{key-measures}}

\hypertarget{music-tastes}{%
\subsubsection{Music tastes}\label{music-taste}}

We operationalize taste using a set of items from the SPPA.
Respondents are asked to provide a binary yes/no response to the question, ``Do you like to listen to \{music genre\}?''
Seventeen music genres are admitted into consideration: (1) classical or chamber music, (2) opera, (3) Broadway musicals or show tunes, (4) jazz, (5) blues or rhythm and blues, (6) gospel or other religious music, (7) country music, (8) bluegrass, (9) folk music, (10) classic rock, (11) alternative or independent rock, (12) pop music, (13) rap or hip-hop, (14) electronic or dance music, (15) reggae, (16) Latin or salsa music, and (17) Asian, African, or Middle-Eastern music.
We treat each affirmative response as an expression of taste for that genre.
For each respondent $i$, we collect these affirmative responses into a taste set $t_{i}$, defined as the subset of the seventeen genres for which respondent $i$ reports a liking.
This set-valued representation of taste serves as the basic unit for the distance measures introduced below, while the genre-level binary indicators serve as the outcomes in the linear probability models described in Section~\ref{positional-fidelity-of-silicon-tastes}.

\hypertarget{genre-popularity}{%
\subsubsection{Genre popularity}\label{genre-popularity}}
We also assign each music genre a binary popularity label.
This label is grounded in data from the \emph{Billboard Hot 100}, the music industry's benchmark weekly ranking of the best-performing songs in the United States.
We compiled every weekly Hot 100 chart from January 2005 through January 2012, a seven-year window closing immediately prior to the SPPA wave we analyze.
Billboard does not itself assign genre, so we matched each pair to the Apple iTunes catalog by programmatic query on title and artist, and recorded the primary genre iTunes assigns to the matched track.
Five genres together account for the large majority of charting songs: pop, rap/hip-hop, country, alternative rock, and electronic or dance music.
Three further genres register a modest but non-trivial chart presence: Asian/African/Middle-Eastern music, reggae, and blues.
For the sake of parsimony, we collapse all eight of these genres into a common ``popular'' category.
The remaining nine genres have effectively no presence on the Hot 100 over this window and labelled as ``not popular.''

\hypertarget{jaccard-distance}{%
\subsubsection{Jaccard distance}\label{jaccard-distance}}

We use a distance measure based on Jaccard similarity (``Jaccard distance'') to measure the differences between synthetic and sample tastes.
Jaccard similarity is a statistical measure used to quantify how similar two sets are by comparing their intersection to their union.
Let $t_{i}$ refer to the set of music genres respondent $i$ expresses likings for.
Let $t_{\hat{\imath}}$ refer to the set of music genres the synthetic counterpart of $i$ would express likings for.
The Jaccard similarity ($J(t_{i}, t_{\hat{\imath}})$) between $t_{i}$ and $t_{\hat{\imath}}$ is given by

\[\frac{|t_{i} \cap t_{\hat{\imath}} |}{|t_{i} \cup t_{\hat{\imath}}|}.\]

\noindent We define Jaccard distance ($D$) simply as $1 - J(t_{i}, t_{\hat{\imath}})$.

\hypertarget{bivariate_relations-between-tastes}{%
\subsubsection{Bivariate relations between tastes}\label{bivariate_relations-between-tastes}}
We consider how well silicon samples preserve the relationality of tastes by measuring the bivariate associations between each pair of genre preferences.
Because the SPPA taste items are binary, we use Cramer's V, a chi-square-based coefficient of association appropriate for categorical variables.
For any two genres $i$ and $j$, we cross-tabulate respondents' genre preferences into a $2\times2$ contingency table and let $\chi^{2}_{ij}$ denote the Pearson chi-square statistic for that table.
Cramer's V is then given by

\[C_{ij} = \sqrt{\frac{\chi^{2}_{ij} / n}{\min(r-1,\,c-1)}},\]

\noindent where $n$ is the number of respondents and $r$ and $c$ are the numbers of rows and columns in the table.
For each sample we compute $C_{ij}$ for every genre pair, which yields a symmetric $17 \times 17$ association matrix whose off-diagonal cells comprise $136$ unique pairs.
We refer to the Cramer's V estimates obtained from the SPPA as $\tilde C_{ij}$, and to those obtained from the silicon and bootstrap samples as $\hat C_{ij}$.

\hypertarget{coefficient-bias}{%
\subsubsection{Coefficient bias}\label{coefficient-bias}}
We use coefficient bias to operationalize the relationship between cultural tastes and social location.
We estimate the coefficient bias of silicon samples in two stages.
First, for each survey wave we fit a linear probability model and extract the coefficients of interest.
Second, we treat these wave-level coefficients as observations in a mixed-effects meta-regression, which lets us pool them while distinguishing sampling error from systematic between-wave variation.
A mixed-effects model assumes that each observed estimate departs from an underlying mean for two reasons: sampling error, and genuine between-wave heterogeneity.
The latter can arise from true differences across waves in populations, historical periods, study designs, analytical methods, or in our case, distinct data-generating procedures.

Let $w \in \{1,\dots,W\}$ index the survey waves generated in the study.\footnotemark \footnotetext{ The $W = 91$ waves comprise the original SPPA data, 30 bootstrap resamples of the 2012 SPPA wave, 30 waves from OpenAI, 15 from DeepSeek, and 15 from Anthropic.}
For each survey wave $w$, we fit a separate linear probability model for each genre $j$, where person $i$'s taste $t$ for genre $j$ is expressed as

\[t_{ij} = \beta_{0j} + \beta_{1j}x_{i1} + \dots + \beta_{7j} x_{i7}.\]

\noindent where the independent variables include the terms for (a) age, (b) age$^2$, (c) income (in US\$1,000s), (d) gender (coded as a binary where 1 = woman), (e) race (coded as a binary where 1 = white), (f) years of education, and (g) urban residence (coded as a binary where 1 = urban).
We are interested in four key estimands from each model: (1) the main effect of age, $\beta_\text{age}$, (2) the main effect of gender, $\beta_\text{gender}$, (3) the main effect of income, $\beta_\text{income}$, and (4) the main effect of race, $\beta_\text{race}$.

We then fit a mixed-effects model to each estimand.
Let $\hat\beta_{\text{age},w}$ be the observed estimate of the age estimand from survey wave $w$, modeled as

\[\hat\beta_{\text{age},w} = \theta_0 + \theta_1z_{w1} +  \dots +  \theta_5z_{w5} + \eta_w + \epsilon_w,\]
\[\text{where } \eta_w \sim N(0, \tau^2) \text{ and } \epsilon_w \sim N(0, \sigma^2_w).\]

\noindent Here $z_{wk}$ denotes the value of the $k$-th moderator variable for wave $w$, while $\eta_w$ and $\epsilon_w$ represent the between-wave heterogeneity and the sampling error of each wave, respectively.
We include the following moderator variables: (a) label for synthetic data, (b) label for popularity of music genre, (c) the interaction between synthetic data and popularity of genre, (d) a label for large magnitude effect and (e) the interaction between synthetic data and the large magnitude effect.
The construction of these labels was discussed above.
We fit the model using the restricted maximum-likelihood estimator (Raudenbush 2009), as implemented by the \emph{metafor} package in R.
We repeat the same procedure for $\beta_\text{gender}$, $\beta_\text{income}$, and $\beta_\text{race}$.

\hypertarget{results}{%
\section{Results}\label{results}}

\hypertarget{ecological-fidelity-of-silicon-tastes}{%
\subsection{Ecological fidelity of silicon tastes}\label{ecological-fidelity-of-silicon-tastes}}

First, we assess the fidelity of ecological estimates produced from silicon samples.
Table~\ref{tab:ecological_estimates} reports ecological estimates of music tastes across the different survey samples.
We find a pervasize positive bias across silicon samples from all LLM providers.
That is to say, silicon surrogates \emph{over-like}, are much more likely to express preferences (``likings'') for music genres than real persons.
This tendency becomes particularly salient when considered in the aggregate.
When looking at an additive index of music tastes\footnotemark \footnotetext{This is the standard measure of omnivorousness used in cultural sociology (e.g. \citealt{peterson_changing_1996}).}, we can observe a clear failure in ecological estimates of taste: silicon respondents look like hyper-omnivores when compared to flesh-and-blood counterparts.
\subsubsection{Pervasive positive bias (over-liking).}
The deltas in Table~\ref{tab:ecological_estimates} are overwhelmingly positive across all three providers.
Silicon respondents over-estimate the share of the population that likes a genre for 10 of the 17 genres under OpenAI, 14 under DeepSeek, and 13 under Anthropic.
The positive deltas are also large in magnitude, reaching $+63.61$ percentage points for folk music (OpenAI), $+62.16$ for gospel (DeepSeek), and $+49.42$ for gospel (Anthropic).
By contrast, under-estimation is both rarer and milder.
Only three genres are under-liked by every provider, and the largest under-estimate in the entire table is just $-9.15$ percentage points (reggae, OpenAI).

\subsubsection{Super-omnivorous, especially in nonpopular genres.}
All of this aggregates into a peculiar variety of omnivorousness among silicon respondents.
If contemporary American tastes are omnivorous (\citealt{peterson_changing_1996,lizardo_reconceptualizing_2012}), then silicon samples of the same may be described as super-omnivorous.
On average, real SPPA respondents report liking $24.18\%$ of the seventeen genres; their silicon counterparts inflate this omnivorousness index by a significant amount ($+21.71\%$ for OpenAI, $+21.96$ for DeepSeek, $+15.59$ for Anthropic).
This tendency for super-omnivorousness is elevated among the non-popular genres of music.
Against a human baseline of $26.52\%$, the mean share of non-popular genres liked rises by $+32.29\%$ (OpenAI), $+29.41\%$ (DeepSeek), and $+20.43\%$ (Anthropic).

\begin{table}[H]

  \caption{Evaluating ecological fidelity}
  \label{tab:ecological_estimates}
  \centering
  \begin{tabular}[t]{lrrrrrrr}
  \toprule
  \multicolumn{2}{c}{ } & \multicolumn{2}{c}{OpenAI} & \multicolumn{2}{c}{DeepSeek} & \multicolumn{2}{c}{Anthropic} \\
  \cmidrule(l{3pt}r{3pt}){3-4} \cmidrule(l{3pt}r{3pt}){5-6} \cmidrule(l{3pt}r{3pt}){7-8}
   & SPPA & Delta & p-val & Delta & p-val & Delta & p-val\\
  \midrule
  \addlinespace[0.3em]
  \multicolumn{8}{l}{\textbf{Popular Genres (\% liked)}}\\
  \hspace{1em}Pop music & 37.09 & 12.97 & 0.00 & 11.48 & 0.00 & 9.05 & 0\\
  \hspace{1em}Hip-hop & 19.02 & -0.66 & 0.25 & 0.32 & 0.58 & 5.22 & 0\\
  \hspace{1em}Country music & 43.62 & 33.56 & 0.00 & 30.10 & 0.00 & 32.94 & 0\\
  \hspace{1em}Alternative rock & 19.82 & -1.51 & 0.01 & 2.90 & 0.00 & 4.45 & 0\\
  \hspace{1em}Electronic music & 15.18 & -4.21 & 0.00 & 0.37 & 0.49 & 4.83 & 0\\
  \hspace{1em}Ethnic music & 8.47 & -0.63 & 0.12 & -3.74 & 0.00 & -3.42 & 0\\
  \hspace{1em}Reggae & 16.09 & -9.15 & 0.00 & -1.76 & 0.00 & -5.12 & 0\\
  \hspace{1em}Blues & 30.94 & 42.66 & 0.00 & 61.49 & 0.00 & 26.64 & 0\\
  \addlinespace[0.3em]
  \multicolumn{8}{l}{\textbf{Non-popular Genres (\% liked)}}\\
  \hspace{1em}Bluegrass & 18.64 & 52.15 & 0.00 & 45.68 & 0.00 & 41.63 & 0\\
  \hspace{1em}Broadway & 22.15 & 19.92 & 0.00 & 29.13 & 0.00 & 13.14 & 0\\
  \hspace{1em}Classical music & 28.21 & 28.25 & 0.00 & 26.07 & 0.00 & 19.26 & 0\\
  \hspace{1em}Classic rock & 51.26 & 44.86 & 0.00 & 43.83 & 0.00 & 47.61 & 0\\
  \hspace{1em}Folk music & 18.76 & 63.61 & 0.00 & 42.13 & 0.00 & 26.50 & 0\\
  \hspace{1em}Gospel & 26.45 & 60.00 & 0.00 & 62.16 & 0.00 & 49.42 & 0\\
  \hspace{1em}Jazz & 27.73 & 37.13 & 0.00 & 29.51 & 0.00 & 3.90 & 0\\
  \hspace{1em}Latin music & 17.83 & -5.44 & 0.00 & -7.46 & 0.00 & -6.53 & 0\\
  \hspace{1em}Opera & 9.83 & -4.38 & 0.00 & 1.14 & 0.01 & -4.56 & 0\\
  \addlinespace[0.3em]
  \multicolumn{8}{l}{\textbf{Aggregate Tastes}}\\
  \hspace{1em}Mean \% of popular genres liked & 23.78 & 9.13 & 0.00 & 12.64 & 0.00 & 9.33 & 0\\
  \hspace{1em}Mean \% of nonpopular genres liked & 26.52 & 32.29 & 0.00 & 29.41 & 0.00 & 20.43 & 0\\
  \hspace{1em}Mean \% of all genres liked & 24.18 & 21.71 & 0.00 & 21.96 & 0.00 & 15.59 & 0\\
  \bottomrule
  \end{tabular}
  \end{table}

\subsubsection{Taste inflation not explained by the minority penalty.}
The minority penalty does not effectively explain the systematic inflation we observed above.
Table~\ref{tab:jaccard_distance_model} reports the coefficient estimates from a linear model of the jaccard distance between silicon samples (pooled) and SPPA data.
The minority penalty hypothesizes that groups whose perspectives and discourse are less represented in the training corpora of large language models -- racial minorities, the less educated, and the less affluent -- should be the hardest to simulate, and thus exhibit the largest Jaccard distances.
But that is not what we observe.
In the \emph{All}-genre model of Table~\ref{tab:jaccard_distance_model}, where a positive coefficient denotes greater distance and therefore \emph{worse} fidelity, three of the four conventional status axes run in the opposite direction.
We find that (a) wealthier respondents are more difficult to simulate, with each additional \$1{,}000 of income raising the Jaccard distance by $0.002$ ($p<0.01$); (b) white respondents are simulated worse, not better, than non-white respondents ($+0.02$, $p<0.01$); and (c) respondents with more education are likewise simulated worse, by $0.01$ per additional year of schooling ($p<0.01$).
Gender is the only axis on which the penalty holds: women are simulated worse than men ($+0.03$, $p<0.01$).

\begin{table}[H] \centering
  \caption{Evaluating WEIRD-bias on ecological fidelity}
  \label{tab:jaccard_distance_model}
\small
\begin{tabular}{@{\extracolsep{1pt}}lccc}
\\[-1.8ex]\hline
\hline \\[-1.8ex]
 & \multicolumn{3}{c}{Jaccard distance} \\
\cline{2-4}
 & All genres & Only popular genres & Only nonpopular genres \\
\hline \\[-1.8ex]
 Age & $-$0.02$^{***}$ & $-$0.04$^{***}$ & 0.003 \\
  & ($-$0.03, $-$0.02) & ($-$0.04, $-$0.03) & ($-$0.003, 0.01) \\
  Age-squared & 0.002$^{***}$ & 0.0002 & 0.003$^{***}$ \\
  & (0.002, 0.003) & ($-$0.0004, 0.001) & (0.002, 0.004) \\
  Income (1,000s) & 0.002$^{***}$ & 0.001$^{***}$ & 0.003$^{***}$ \\
  & (0.002, 0.003) & (0.001, 0.001) & (0.003, 0.004) \\
  Education (years) & 0.01$^{***}$ & 0.003$^{***}$ & 0.01$^{***}$ \\
  & (0.01, 0.01) & (0.003, 0.004) & (0.01, 0.01) \\
  Gender (woman) & 0.03$^{***}$ & 0.01$^{***}$ & 0.05$^{***}$ \\
  & (0.03, 0.03) & (0.003, 0.01) & (0.05, 0.06) \\
  Race (white) & 0.02$^{***}$ & $-$0.06$^{***}$ & 0.09$^{***}$ \\
  & (0.01, 0.02) & ($-$0.07, $-$0.06) & (0.08, 0.10) \\
  Urban residence & 0.01$^{***}$ & 0.07$^{***}$ & $-$0.04$^{***}$ \\
  & (0.004, 0.01) & (0.07, 0.08) & ($-$0.04, $-$0.03) \\
  Constant & 0.26$^{***}$ & 0.47$^{***}$ & 0.07$^{***}$ \\
  & (0.25, 0.28) & (0.45, 0.48) & (0.06, 0.09) \\
 \hline \\[-1.8ex]
Observations & 27,747 & 27,747 & 27,747 \\
R$^{2}$ & 0.06 & 0.21 & 0.20 \\
Adjusted R$^{2}$ & 0.06 & 0.21 & 0.20 \\
\hline
\hline \\[-1.8ex]
\textit{Note:}  & \multicolumn{3}{r}{$^{*}$p$<$0.1; $^{**}$p$<$0.05; $^{***}$p$<$0.01} \\
\end{tabular}
\end{table}

\hypertarget{relationality-of-tastes}{%
\subsection{Relationality of tastes}\label{relationality-of-tastes}}

Second, we investigate the relationality of silicon tastes by looking at the association matrices of tastes.
Human tastes have a rich and persistent relational structure, to which silicon tastes hold little in common.
For each sample we compute the $17 \times 17$ matrix of pairwise Cramer's V associations introduced in Section~\ref{bivariate_relations-between-tastes}, writing the SPPA estimates as $\tilde C_{ij}$ and those from the silicon and bootstrap samples as $\hat C_{ij}$.
Table~\ref{tab:cramersv_silicon_pooled} reports this association matrix for the silicon samples, with all providers pooled together.
Figure~\ref{scatter_cramerv} plots each genre pair's sample Cramer's V ($\hat C_{ij}$, y-axis) against its Cramer's V in the SPPA ($\tilde C_{ij}$, x-axis).\footnotemark \footnotetext{A point falling on the dashed $45^{\circ}$ line denotes a pair whose association is reproduced exactly; points above the line are coupled more strongly in the sample than among human respondents, and points below it more weakly. The bootstrap panel serves as a sanity check.}
Figure~\ref{heatmap_cramerv} plots the cell-wise differences ($\hat C_{ij} - \tilde C_{ij}$) as a grid of four heat maps, one for each sample source.
Appendix~\ref{appendix-cramerv} contains a more detailed breakdown of association matrices by each of the LLM providers.

\subsubsection{The relational structure of silicon tastes is misaligned with human tastes.}
Across all $136$ unique genre pairs (Table~\ref{tab:cramersv_silicon_pooled}), the pairwise Cramer's V values from the silicon samples are essentially uncorrelated with those from the SPPA (Pearson's $r = 0.06, p < 0.01$).
From Figure~\ref{scatter_cramerv}, we can also see obvious differences between the spread of the silicon samples and the true estimates.
Cramer's V estimates from silicon samples form a diffuse cloud bearing little relation to the diagonal.

\begin{figure}[H]
  \centering
  \caption{Heat map of Deviation in Cramer's V}\label{heatmap_cramerv}
  \includegraphics[width = \linewidth]{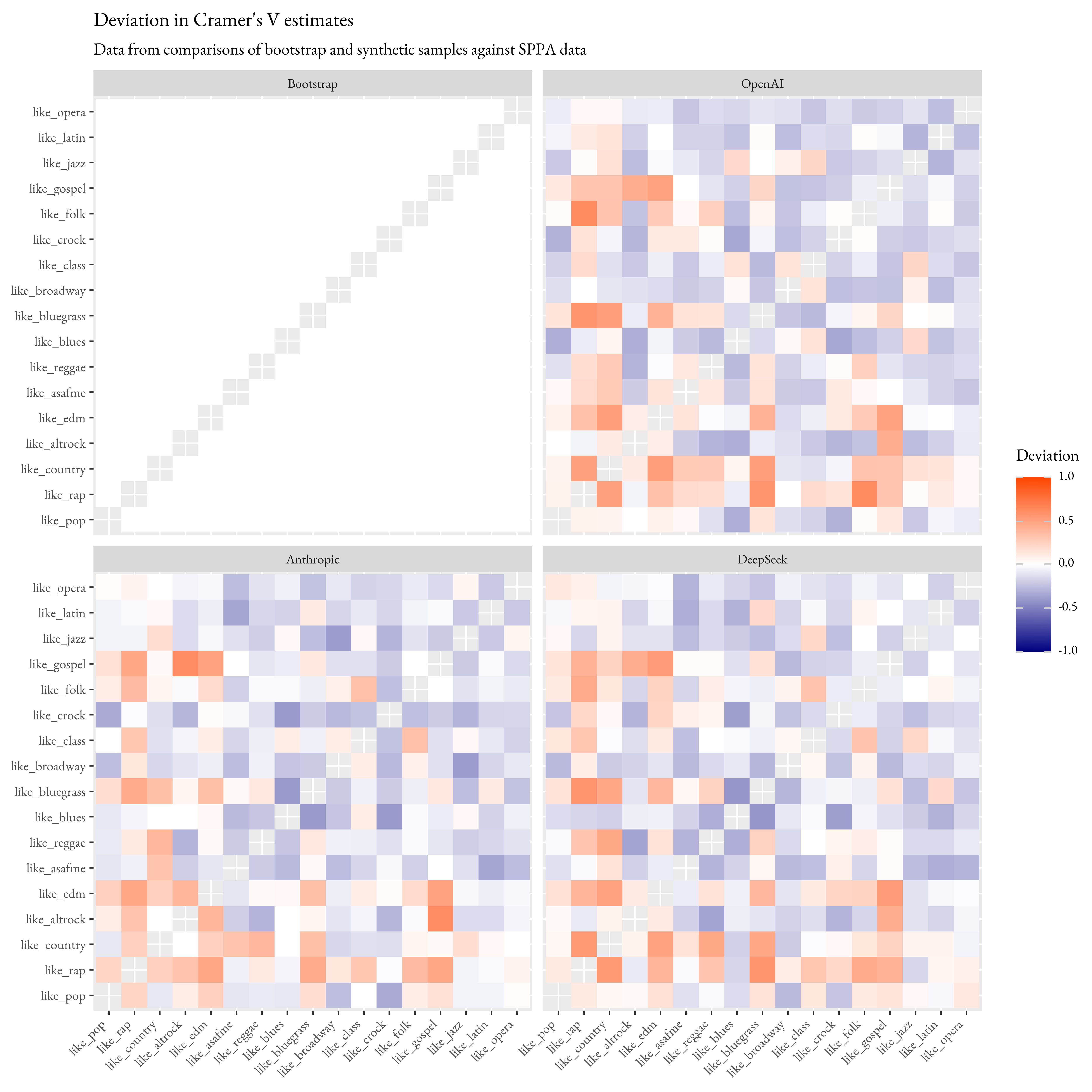}
\end{figure}

For a more substantive example of this misalignment, consider how silicon tastes misrepresent the preference associations between music genres that are best known.
Comparing the SPPA matrix (Table~\ref{tab:cramersv_sppa}) with the pooled silicon matrix (Table~\ref{tab:cramersv_silicon_pooled}), we find that several of the strongest human associations among these genres are sharply attenuated: the blues/classic rock association falls from $0.40$ to $0.04$, blues/bluegrass from $0.44$ to $0.16$, classical/opera from $0.44$ to $0.26$, and Broadway/gospel from $0.32$ to $0.10$.
These pairs show up as cool (blue) cells in the LLM panels of Figure~\ref{heatmap_cramerv}.
Conversely, silicon manufactures associations among mainstream genres that humans do not actually link (visible as warm (orange) cells in Figure~\ref{heatmap_cramerv}).
Relative to the SPPA, the silicon samples inflate the country/bluegrass association from $0.33$ to $0.80$, electronic/hip-hop from $0.34$ to $0.72$, hip-hop/bluegrass from $0.07$ to $0.61$, folk/hip-hop from $0.07$ to $0.59$, and electronic/gospel from $0.06$ to $0.57$.

\subsubsection{Silicon tastes have ersatz properties.}
However, when looking at aggregate measures of relationality, silicon tastes can appear deceivingly similar to real tastes.
The mean Cramer's V across all genre-pairs is $0.247$ for the pooled silicon samples (full tablecan be found in Table~\ref{tab:cramersv_silicon_pooled}), only marginally below the $0.264$ observed in the SPPA.
At this coarsest level of analysis, silicon taste therefore looks about as ``relational'' as human taste.
Yet as we document above, the two sets of taste differ in important ways.
Their relationality is in this sense ersatz, reproducing a superficial facsimile of cultural tastes that severely distorts its relational structure.

\subsubsection{Relationality fidelity is conditional on popularity; worse for less popular genres.}
\begin{figure}[H]
  \centering
  \caption{Comparing Cramer's V estimates of Silicon Samples against SPPA data}\label{scatter_cramerv}
  \includegraphics[width = \linewidth]{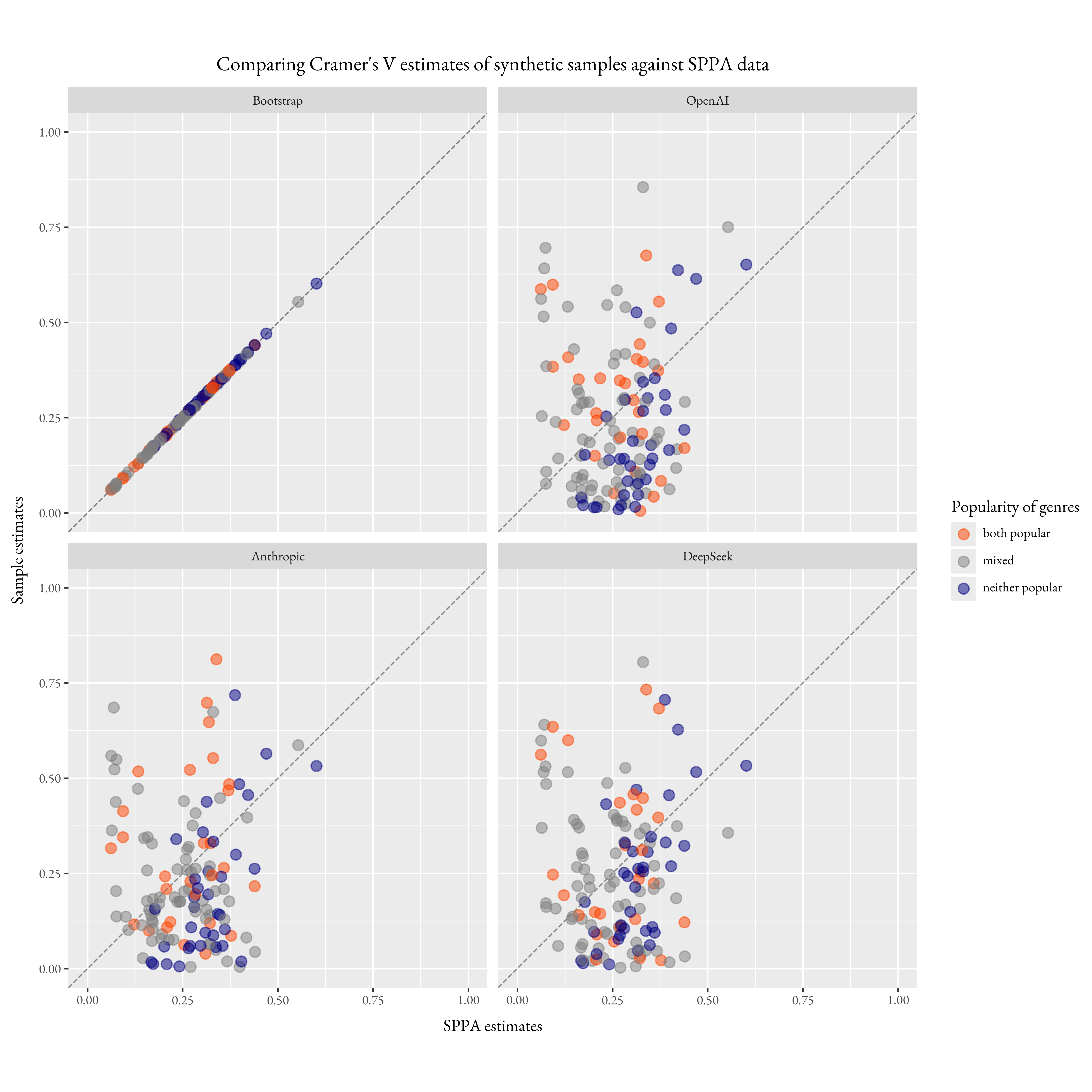}
\end{figure}

Further, we find that the variance in relational fidelity is conditional on the popularity of the genres involved.
The coloring of Figure~\ref{scatter_cramerv} shows that the displacement from the diagonal is patterned by popularity: pairs in which both genres are popular (orange) sit disproportionately above the diagonal, pairs in which neither is popular (blue) fall below it, and mixed pairs (grey) scatter around it.
Silicon samples exaggerate the associatons between mainstream genres and attenuate the ones between less popular genres.
The average deviation in Cramer's V is $+0.04$ for both-popular pairs, approximately zero for mixed pairs, and $-0.10$ for neither-popular pairs.

\begin{landscape}
  \begin{table}
    \caption{Cramer's V between music genres (silicon samples, all providers pooled)}
    \label{tab:cramersv_silicon_pooled}
    \centering
    \footnotesize
    \begin{tabular}[t]{lrrrrrrrrrrrrrrrrr}
    \toprule
    & \rotatebox{90}{Pop music} & \rotatebox{90}{Hip-hop} & \rotatebox{90}{Country} & \rotatebox{90}{Alternative rock} & \rotatebox{90}{Electronic} & \rotatebox{90}{Ethnic music} & \rotatebox{90}{Reggae} & \rotatebox{90}{Blues} & \rotatebox{90}{Bluegrass} & \rotatebox{90}{Broadway} & \rotatebox{90}{Classical} & \rotatebox{90}{Classic rock} & \rotatebox{90}{Folk} & \rotatebox{90}{Gospel} & \rotatebox{90}{Jazz} & \rotatebox{90}{Latin} & \rotatebox{90}{Opera} \\
    \midrule
    Pop music &  & 0.45 & 0.30 & 0.40 & 0.41 & 0.17 & 0.24 & 0.14 & 0.41 & 0.12 & 0.20 & 0.13 & 0.34 & 0.30 & 0.21 & 0.21 & 0.17 \\
    Hip-hop & 0.45 &  & 0.54 & 0.35 & 0.72 & 0.24 & 0.57 & 0.18 & 0.61 & 0.13 & 0.31 & 0.29 & 0.59 & 0.45 & 0.13 & 0.25 & 0.13 \\
    Country & 0.30 & 0.54 &  & 0.19 & 0.51 & 0.36 & 0.48 & 0.19 & 0.80 & 0.09 & 0.05 & 0.29 & 0.47 & 0.46 & 0.30 & 0.19 & 0.11 \\
    Alternative rock & 0.40 & 0.35 & 0.19 &  & 0.48 & 0.06 & 0.07 & 0.15 & 0.22 & 0.06 & 0.08 & 0.05 & 0.11 & 0.56 & 0.09 & 0.05 & 0.11 \\
    Electronic & 0.41 & 0.72 & 0.51 & 0.48 &  & 0.24 & 0.34 & 0.17 & 0.52 & 0.06 & 0.18 & 0.27 & 0.40 & 0.57 & 0.12 & 0.22 & 0.10 \\
    Ethnic music & 0.17 & 0.24 & 0.36 & 0.06 & 0.24 &  & 0.26 & 0.10 & 0.35 & 0.03 & 0.10 & 0.25 & 0.25 & 0.19 & 0.16 & 0.11 & 0.07 \\
    Reggae & 0.24 & 0.57 & 0.48 & 0.07 & 0.34 & 0.26 &  & 0.17 & 0.44 & 0.10 & 0.19 & 0.27 & 0.43 & 0.09 & 0.21 & 0.20 & 0.08 \\
    Blues & 0.14 & 0.18 & 0.19 & 0.15 & 0.17 & 0.10 & 0.17 &  & 0.16 & 0.29 & 0.44 & 0.04 & 0.28 & 0.18 & 0.61 & 0.07 & 0.14 \\
    Bluegrass & 0.41 & 0.61 & 0.80 & 0.22 & 0.52 & 0.35 & 0.44 & 0.16 &  & 0.11 & 0.15 & 0.26 & 0.59 & 0.49 & 0.23 & 0.32 & 0.07 \\
    Broadway & 0.12 & 0.13 & 0.09 & 0.06 & 0.06 & 0.03 & 0.10 & 0.29 & 0.11 &  & 0.58 & 0.08 & 0.32 & 0.10 & 0.31 & 0.06 & 0.29 \\
    Classical & 0.20 & 0.31 & 0.05 & 0.08 & 0.18 & 0.10 & 0.19 & 0.44 & 0.15 & 0.58 &  & 0.11 & 0.51 & 0.09 & 0.59 & 0.18 & 0.26 \\
    Classic rock & 0.13 & 0.29 & 0.29 & 0.05 & 0.27 & 0.25 & 0.27 & 0.04 & 0.26 & 0.08 & 0.11 &  & 0.26 & 0.02 & 0.11 & 0.02 & 0.03 \\
    Folk & 0.34 & 0.59 & 0.47 & 0.11 & 0.40 & 0.25 & 0.43 & 0.28 & 0.59 & 0.32 & 0.51 & 0.26 &  & 0.28 & 0.24 & 0.29 & 0.15 \\
    Gospel & 0.30 & 0.45 & 0.46 & 0.56 & 0.57 & 0.19 & 0.09 & 0.18 & 0.49 & 0.10 & 0.09 & 0.02 & 0.28 &  & 0.11 & 0.16 & 0.05 \\
    Jazz & 0.21 & 0.13 & 0.30 & 0.09 & 0.12 & 0.16 & 0.21 & 0.61 & 0.23 & 0.31 & 0.59 & 0.11 & 0.24 & 0.11 &  & 0.09 & 0.26 \\
    Latin & 0.21 & 0.25 & 0.19 & 0.05 & 0.22 & 0.11 & 0.20 & 0.07 & 0.32 & 0.06 & 0.18 & 0.02 & 0.29 & 0.16 & 0.09 &  & 0.04 \\
    Opera & 0.17 & 0.13 & 0.11 & 0.11 & 0.10 & 0.07 & 0.08 & 0.14 & 0.07 & 0.29 & 0.26 & 0.03 & 0.15 & 0.05 & 0.26 & 0.04 &  \\
    \bottomrule
    \end{tabular}
  \end{table}
  \end{landscape}

\hypertarget{positional-fidelity-of-silicon-tastes}{%
\subsection{Positional fidelity of silicon tastes}\label{positional-fidelity-of-silicon-tastes}}

Finally, we consider how closely silicon samples are able to reproduce well-understood associations between tastes and social positions.
Table ~\ref{tab:coefbias_combined} contains results from our meta-regression models.
We include original coefficient estimates from SPPA data, and the expected bias introduced by silicon samples.
As mentioned above, we restrict ourselves to four major dimensions of social space: age, income, gender, and race.
We discuss each in turn.

\subsubsection{Silicon samples juvenilize musical taste.}
In silicon samples, age and tastes for music are much more likely to share a negative association with each other.
In Table~\ref{tab:coefbias_combined}, the age-bias column is negative for 12 of the 17 genres.
To pick some specific examples, alternative rock carries a modest true age effect of $-0.025$ in the SPPA, yet picks up a bias of $-0.504$, roughly a twenty-fold inflation in the age effect.
Hip-hop, already the most youth-coded genre in the human data ($-0.193$), is pushed further in the same direction ($-0.102$ bias).
Age-taste associations are also clearly distorted for classic rock.
Among human respondents it is strongly elder-coded ($+0.201$), but among silicon samples this is attenuated significantly ($-0.119$).
However, there are important exceptions to this general trend.
The age/taste associations for country and bluegrass are faithfully reproduced by silicon samples.
On the other hand, silicon samples exaggerate the senscence of a small set of music genres.
Blues ($+0.121$ true, $+0.150$ bias) and gospel ($+0.063$ true, $+0.131$ bias) both pick up positive bias that pushes their age effects further into older territory.

\subsubsection{Silicon samples resurrect retrograde class-taste homologies.}
Consistent with broad changes in American taste structures over the past half-century (\citealt{bryson_anything_1996, peterson_changing_1996,lizardo_reconceptualizing_2012}), we find that income and tastes to be relatively decoupled with each other in the SPPA data.\footnote{Notably, logged income does not have any linear association with tastes for the canonical ``high-brow'' genres like classical ($+0.006$, $p=0.20$), opera ($-0.004$, $p=0.28$), jazz ($+0.007$, $p=0.19$), and blues ($-0.004$, $p=0.49$).}
By sharp contrast, silicon tastes are retrograde.
Among silicon tastes, we observe ight coupling between class and taste that is more redolent of mid 20th century high-low taste hierarchies (\citealt{bourdieu_distinction:_1984, levine_highbrow/lowbrow:_1988}).
The income-bias column is positive and statistically significant for 11 of the 17 genres.
The largest income-biases are observed for the ``highbrow'' genres: classical ($+0.096$ bias against a true effect of $+0.006$), jazz ($+0.078$ against $+0.007$), blues ($+0.067$ against true effects of $-0.004$), folk ($+0.027$ against $-0.007$), and opera ($+0.025$ bias, against $-0.004$ true).

\subsubsection{Silicon samples caricaturize gender-taste associations}
The SPPA data does contain real gender-taste associations, but they are modest and spread across several genres.
Among silicon samples, a smaller set of genres retain these gender-taste associations.
Further, these associations are more exaggerated than they are in the SPPA sample.
In this sense, gender-taste associations are caricaturized.
In Table~\ref{tab:coefbias_combined}, eight genres register significant gender (woman) coefficients in the SPPA: Broadway ($+0.107$), gospel ($+0.097$), pop ($+0.083$), country ($+0.054$), Latin ($+0.039$), classical ($+0.032$), blues ($+0.023$), and opera ($+0.023$).
Hip-hop ($+0.015$) and reggae ($+0.015$) show small positive effects as well.
Silicon samples caricaturize these gender-taste associations by concentrating them into a smaller set of genres.
The three largest gender biases in the table fall on Broadway ($+0.453$), pop ($+0.328$), and classical ($+0.125$), each roughly three- to four-times the corresponding SPPA coefficient.
Folk music also picks up a bias of $+0.082$ against a true effect of essentially zero ($+0.003$), manufacturing a gender gradient where the SPPA had shown none.
By contrast, silicon attenuates or reverses associations that complicate the stereotype.
Country is the clearest case: women genuinely like country music more than men in the human data ($+0.054$), yet the silicon bias is negative ($-0.046$), erasing one of the strongest non-stereotypical gender effects in the table.
Hip-hop and reggae show the same sign problem in miniature: both are slightly women-leaning in the SPPA ($+0.015$ each), but their biases run negative ($-0.031$ and $-0.036$), pushing silicon samples toward a male-coded account of genres that human women in fact report liking at least as much as men.

\subsubsection{Fictive race-taste associations.}
Race-taste associations are where silicon samples show the poorest fidelity.
In Table~\ref{tab:coefbias_combined}, the largest race (white) coefficients in the SPPA are country ($+0.202$) and gospel ($-0.129$); no other genre exceeds $+0.183$ (classic rock) or falls below $-0.117$ (blues).
Silicon samples push the implied coefficients far beyond this range, from $+0.50$ for bluegrass ($+0.122 + +0.377$) and $+0.48$ for folk ($+0.099 + +0.384$), to $-0.43$ for hip-hop ($-0.110 + -0.320$), producing an exaggerated racialization music taste that far exceeds anything in the SPPA data.
We can decompose this racialization into two components, (a) the exaggeration of mild race-taste association and (b) the introduction of wholly fictive race-taste associations.
Silicon samples exaggerate mild biases for certain race-taste associations, such as folk ($+0.384$ on a true effect of $+0.099$), bluegrass ($+0.377$ on $+0.122$), and country ($+0.229$ on $+0.202$).
Broadway and classical music receive this on a smaller scale.
Among genres that are registered as nonwhite-leaning in the SPPA, the amplification runs the other way: reggae ($-0.365$ on $-0.039$, roughly a nine-fold inflation and the most extreme entry in the table), jazz ($-0.331$ on $-0.072$), hip-hop ($-0.320$ on $-0.110$), and ethnic music ($-0.284$ on $-0.080$).
Second, silicon samples can also invent race-taste associations.
Electronic music is one example, where the race-taste coefficient is null under SPPA data, yet silicon samples assigns a bias of $-0.152$, fabricating a non-white association from nothing.
Pop and gospel music provide the best demonstrations of this.
Pop music is white-leaning in the SPPA ($+0.075$), but its race bias is $-0.240$, reversing the sign of the implied silicon coefficient to roughly $-0.17$.
On the other hand, silicon samples assume white audiences to prefer gospel music much more than they actually do.
Gospel is one of the most non-white coded genres in the SPPA data ($-0.129$), but this association is severely attenuated in the data ($+0.068$).

\begin{landscape}
\begin{table}
  \centering
  \caption{Estimated bias in coefficient estimates from silicon samples: age, logged income, gender, and race}
  \label{tab:coefbias_combined}
  \footnotesize
  \setlength{\tabcolsep}{4pt}
  \resizebox{\linewidth}{!}{%
  \begin{tabular}[t]{l rrrr rrrr rrrr rrrr}
  \toprule
   & \multicolumn{4}{c}{Age} & \multicolumn{4}{c}{Logged income} & \multicolumn{4}{c}{Gender (woman)} & \multicolumn{4}{c}{Race (white)} \\
  \cmidrule(l{3pt}r{3pt}){2-5} \cmidrule(l{3pt}r{3pt}){6-9} \cmidrule(l{3pt}r{3pt}){10-13} \cmidrule(l{3pt}r{3pt}){14-17}
   & \multicolumn{2}{c}{Bias} & \multicolumn{2}{c}{Coef. est.} & \multicolumn{2}{c}{Bias} & \multicolumn{2}{c}{Coef. est.} & \multicolumn{2}{c}{Bias} & \multicolumn{2}{c}{Coef. est.} & \multicolumn{2}{c}{Bias} & \multicolumn{2}{c}{Coef. est.} \\
  \cmidrule(l{2pt}r{2pt}){2-3}\cmidrule(l{2pt}r{2pt}){4-5}\cmidrule(l{2pt}r{2pt}){6-7}\cmidrule(l{2pt}r{2pt}){8-9}\cmidrule(l{2pt}r{2pt}){10-11}\cmidrule(l{2pt}r{2pt}){12-13}\cmidrule(l{2pt}r{2pt}){14-15}\cmidrule(l{2pt}r{2pt}){16-17}
   & Est. & $p$ & Est. & $p$ & Est. & $p$ & Est. & $p$ & Est. & $p$ & Est. & $p$ & Est. & $p$ & Est. & $p$ \\
  \midrule
  \multicolumn{8}{l}{\textbf{Popular Genres}}\\
  \hspace{1em}Pop music & -0.183 & $<$0.01 & 0.001 & 0.94 & 0.016 & $<$0.01 & 0.030 & $<$0.01 & 0.328 & $<$0.01 & 0.083 & $<$0.01 & -0.240 & $<$0.01 & 0.075 & $<$0.01 \\
  \hspace{1em}Hip-hop & -0.102 & $<$0.01 & -0.193 & $<$0.01 & -0.004 & $<$0.01 & -0.005 & 0.20 & -0.031 & $<$0.01 & 0.015 & 0.04 & -0.320 & $<$0.01 & -0.110 & $<$0.01 \\
  \hspace{1em}Country & -0.001 & 0.70 & 0.062 & $<$0.01 & -0.007 & $<$0.01 & -0.003 & 0.64 & -0.046 & $<$0.01 & 0.054 & $<$0.01 & 0.229 & $<$0.01 & 0.202 & $<$0.01 \\
  \hspace{1em}Alternative rock & -0.504 & $<$0.01 & -0.025 & 0.04 & 0.028 & $<$0.01 & 0.008 & 0.09 & -0.002 & 0.21 & -0.011 & 0.15 & 0.013 & $<$0.01 & 0.111 & $<$0.01 \\
  \hspace{1em}Electronic & -0.279 & $<$0.01 & -0.067 & $<$0.01 & 0.014 & $<$0.01 & 0.000 & 0.95 & 0.000 & 0.99 & 0.010 & 0.16 & -0.152 & $<$0.01 & 0.016 & 0.12 \\
  \hspace{1em}Ethnic music & -0.034 & $<$0.01 & 0.019 & 0.03 & 0.021 & $<$0.01 & -0.006 & 0.05 & 0.000 & 0.98 & -0.003 & 0.61 & -0.284 & $<$0.01 & -0.080 & $<$0.01 \\
  \multicolumn{8}{l}{\textbf{Non-popular Genres}}\\
  \hspace{1em}Reggae & -0.038 & $<$0.01 & 0.044 & $<$0.01 & -0.001 & 0.58 & -0.001 & 0.83 & -0.036 & $<$0.01 & 0.015 & 0.04 & -0.365 & $<$0.01 & -0.039 & $<$0.01 \\
  \hspace{1em}Blues & 0.150 & $<$0.01 & 0.121 & $<$0.01 & 0.067 & $<$0.01 & -0.004 & 0.49 & -0.005 & 0.58 & 0.023 & 0.02 & -0.069 & $<$0.01 & -0.117 & $<$0.01 \\
  \hspace{1em}Bluegrass & -0.001 & 0.95 & 0.077 & $<$0.01 & 0.000 & 0.96 & -0.009 & 0.05 & -0.029 & $<$0.01 & -0.009 & 0.24 & 0.377 & $<$0.01 & 0.122 & $<$0.01 \\
  \hspace{1em}Broadway & -0.036 & $<$0.01 & -0.015 & 0.24 & 0.067 & $<$0.01 & 0.015 & $<$0.01 & 0.453 & $<$0.01 & 0.107 & $<$0.01 & 0.096 & $<$0.01 & 0.059 & $<$0.01 \\
  \hspace{1em}Classical & 0.016 & $<$0.01 & 0.018 & 0.20 & 0.096 & $<$0.01 & 0.006 & 0.20 & 0.125 & $<$0.01 & 0.032 & $<$0.01 & 0.058 & $<$0.01 & 0.061 & $<$0.01 \\
  \hspace{1em}Classic rock & -0.119 & $<$0.01 & 0.201 & $<$0.01 & -0.005 & 0.01 & 0.016 & $<$0.01 & -0.014 & $<$0.01 & 0.003 & 0.74 & -0.043 & $<$0.01 & 0.183 & $<$0.01 \\
  \hspace{1em}Folk & -0.014 & $<$0.01 & 0.061 & $<$0.01 & 0.027 & $<$0.01 & -0.007 & 0.09 & 0.082 & $<$0.01 & 0.003 & 0.72 & 0.384 & $<$0.01 & 0.099 & $<$0.01 \\
  \hspace{1em}Gospel & 0.131 & $<$0.01 & 0.063 & $<$0.01 & -0.011 & $<$0.01 & -0.015 & $<$0.01 & -0.003 & 0.45 & 0.097 & $<$0.01 & 0.068 & $<$0.01 & -0.129 & $<$0.01 \\
  \hspace{1em}Jazz & -0.051 & $<$0.01 & 0.098 & $<$0.01 & 0.078 & $<$0.01 & 0.007 & 0.19 & -0.015 & 0.44 & 0.006 & 0.52 & -0.331 & $<$0.01 & -0.072 & $<$0.01 \\
  \hspace{1em}Latin & -0.032 & $<$0.01 & 0.047 & $<$0.01 & 0.003 & $<$0.01 & -0.003 & 0.50 & -0.015 & $<$0.01 & 0.039 & $<$0.01 & -0.023 & $<$0.01 & 0.063 & $<$0.01 \\
  \hspace{1em}Opera & -0.070 & $<$0.01 & -0.008 & 0.37 & 0.025 & $<$0.01 & -0.004 & 0.28 & 0.011 & $<$0.01 & 0.023 & $<$0.01 & -0.025 & $<$0.01 & 0.024 & $<$0.01 \\
  \bottomrule
  \end{tabular}%
  }
\end{table}
\end{landscape}

\pagebreak
\hypertarget{conclusion}{%
\section{Conclusion}\label{conclusion}}

Silicon sampling promises cheap, scalable surrogates for human survey respondents.
This paper asks how far that promise extends to cultural taste, a domain in which scholars practitioners are interested in not just population-level estimates of tastes and preferences, but also their relational structure and alignment with social space more broadly.
Using 554,940 silicon surrogates from OpenAI, Anthropic, and DeepSeek, matched to respondents from the \emph{Survey of Public Participation in the Arts}, we find that LLM-generated ``silicon tastes'' to be stylized facsimiles of human tastes.
While they share ersatz resemblance when looking at certain aggregate measures of proportionality or covariance, they remain systematically distinct from human data in ways that would mislead researchers who reify them as interchangeable.
Synthetic surrogates, it turns out, are prodigious enjoyers of music.
At the ecological level, silicon samples are super-omnivore who over-like, and this is particularly so when it comes to less popular genres of music.
The relational structure of silicon tastes are fundamentally alien to human tastes, and this (lack of) relational fidelity only worsens for less-popular tastes.
Silicon tastes also confound well-understood associations between tastes and social positions.
They juvenilize taste, resurrect class-taste homologies, caricaturize gender-taste associations, and invent fictive race-taste associations.

Our work carry important implications for those interested in the application and development of AI-augmented research methods in the sociology of culture.
First, silicon samples aren't good enough \emph{yet} for applied research on cultural tastes.
As documented above, there are serious concerns around the ecological, relational, and positional fidelity of silicon samples of cultural taste.
We echo previous work practicing caution around their use in research (\citealt{bisbee_synthetic_2024}).
What has changed in recent times, however, is that silicon sampling is no longer just an experimental methodology studied by scholars interested in AI-augmented research (\citealt{bail_can_2024}).
``Synthetic'' survey panels build atop silicon sampling are already being aggresively pushed out by survey and market research companies (\citealt{isaza_can_2024, currie_qualtrics_2025, verasight_can_2026}), such that it is highly likely that scholars and practitioners may encounter and consider their use in the coming months and years.
We urge researchers to be chary with such data.
With all that being said, we stress that we are not outright skeptics of silicon sampling.
Silicon sampling is a technique in its infancy, and these defeciencies may not neccessarily stay that way.
As methodological improvements in model architecture, reinforcement-learning, retrieval-augmented generation, and agentic frameworks occur, re-assessments of the algorithimic fidelity of silicon tastes will be needed.

Second, our study sheds further light on the tastes of the large-language models themselves.
Silicon samples and silicon tastes may be interesting subjects of study for sociologists of culture, regardless of their algorithmic fidelity.
Indeed, given the rapid proliferation of AI agents as both producers and consumers of cultural content (\citealt{brinkmann_machine_2023}), what these models take taste to be is not merely a methodological nuisance to be corrected, but a substantive object of inquiry in its own right.
Read this way, our results describe a distinct counterfactual `world' rather than a failure to be faithful to reality.
It is an odd world in which nearly everyone likes nearly everything; in which tastes quickly ossify as we grow old; in which class sorts music genres into a status hierarchy that American audiences largely abandoned decades ago; and one in which race and gender predict taste far better than we would expect.
Why are silicon tastes so?
Our design cannot adjudicate among mechanisms, but three conjectures seem worth stating.
One, modern large-language models use reinforcement learning from human feedback (RLHF) techniques to better align models with human preferences and values.
Such post-training procedures that reward helpfulness and agreeableness plausibly inflate assent quite generally, and would yield precisely the indiscriminate liking we observe.
Two, the training corpus.
Internet text is dense with discourse \emph{about} culture but comparatively sparse in anything resembling a representative joint distribution of non-discursive cultural behavior.
A model trained on such text would learn that opera is elite and that hip-hop is young, because these are the propositions the corpus asserts, while remaining ignorant to the subtler multivariate patterns that a nationally representative survey recovers.
If this is so, this carries implications for silicon sampling beyond music, and extends to any domain in which discursive and non-discursive modes bear consideration (\citealt{lizardo_cultural_2016}).
This may explain why silicon samples are far better at reproducing aggregate political opinions than cultural preferences (e.g. \citealt{argyle_out_2023}): fidelity should decline as one moves from domains of explicit discourse toward domains of embodied practice, such as that of music taste.

\bibliographystyle{ACM-Reference-Format}
\bibliography{references}

\hypertarget{appendix-cramerv}{%
\section{Appendix: Cramer's V matrices}\label{appendix-cramerv}}

\begin{landscape}
  \begin{table}
    \caption{Cramer's V between music genres (SPPA)}
    \label{tab:cramersv_sppa}
    \centering
    \footnotesize
    \begin{tabular}[t]{lrrrrrrrrrrrrrrrrr}
    \toprule
    & \rotatebox{90}{Pop music} & \rotatebox{90}{Hip-hop} & \rotatebox{90}{Country} & \rotatebox{90}{Alternative rock} & \rotatebox{90}{Electronic} & \rotatebox{90}{Ethnic music} & \rotatebox{90}{Reggae} & \rotatebox{90}{Blues} & \rotatebox{90}{Bluegrass} & \rotatebox{90}{Broadway} & \rotatebox{90}{Classical} & \rotatebox{90}{Classic rock} & \rotatebox{90}{Folk} & \rotatebox{90}{Gospel} & \rotatebox{90}{Jazz} & \rotatebox{90}{Latin} & \rotatebox{90}{Opera} \\
    \midrule
    Pop music &  & 0.33 & 0.28 & 0.37 & 0.27 & 0.21 & 0.33 & 0.36 & 0.25 & 0.32 & 0.26 & 0.42 & 0.28 & 0.17 & 0.32 & 0.25 & 0.17 \\
    Hip-hop & 0.33 &  & 0.09 & 0.32 & 0.34 & 0.16 & 0.37 & 0.27 & 0.07 & 0.08 & 0.06 & 0.16 & 0.07 & 0.08 & 0.17 & 0.19 & 0.08 \\
    Country & 0.28 & 0.09 &  & 0.12 & 0.06 & 0.09 & 0.13 & 0.21 & 0.33 & 0.23 & 0.15 & 0.34 & 0.26 & 0.24 & 0.16 & 0.10 & 0.11 \\
    Alternative rock & 0.37 & 0.32 & 0.12 &  & 0.31 & 0.25 & 0.38 & 0.32 & 0.27 & 0.20 & 0.23 & 0.34 & 0.27 & 0.07 & 0.28 & 0.21 & 0.17 \\
    Electronic & 0.27 & 0.34 & 0.06 & 0.31 &  & 0.22 & 0.31 & 0.20 & 0.13 & 0.17 & 0.16 & 0.16 & 0.15 & 0.06 & 0.17 & 0.24 & 0.14 \\
    Ethnic music & 0.21 & 0.16 & 0.09 & 0.25 & 0.22 &  & 0.32 & 0.31 & 0.26 & 0.27 & 0.31 & 0.18 & 0.32 & 0.19 & 0.30 & 0.37 & 0.32 \\
    Reggae & 0.33 & 0.37 & 0.13 & 0.38 & 0.31 & 0.32 &  & 0.44 & 0.28 & 0.24 & 0.24 & 0.28 & 0.28 & 0.17 & 0.37 & 0.36 & 0.19 \\
    Blues & 0.36 & 0.27 & 0.21 & 0.32 & 0.20 & 0.31 & 0.44 &  & 0.44 & 0.36 & 0.35 & 0.40 & 0.42 & 0.32 & 0.55 & 0.31 & 0.27 \\
    Bluegrass & 0.25 & 0.07 & 0.33 & 0.27 & 0.13 & 0.26 & 0.28 & 0.44 &  & 0.35 & 0.32 & 0.34 & 0.60 & 0.31 & 0.36 & 0.23 & 0.24 \\
    Broadway & 0.32 & 0.08 & 0.23 & 0.20 & 0.17 & 0.27 & 0.24 & 0.36 & 0.35 &  & 0.47 & 0.34 & 0.40 & 0.32 & 0.40 & 0.27 & 0.39 \\
    Classical & 0.26 & 0.06 & 0.15 & 0.23 & 0.16 & 0.31 & 0.24 & 0.35 & 0.32 & 0.47 &  & 0.30 & 0.39 & 0.28 & 0.42 & 0.28 & 0.44 \\
    Classic rock & 0.42 & 0.16 & 0.34 & 0.34 & 0.16 & 0.18 & 0.28 & 0.40 & 0.34 & 0.34 & 0.30 &  & 0.33 & 0.21 & 0.35 & 0.17 & 0.17 \\
    Folk & 0.28 & 0.07 & 0.26 & 0.27 & 0.15 & 0.32 & 0.28 & 0.42 & 0.60 & 0.40 & 0.39 & 0.33 &  & 0.33 & 0.35 & 0.28 & 0.29 \\
    Gospel & 0.17 & 0.08 & 0.24 & 0.07 & 0.06 & 0.19 & 0.17 & 0.32 & 0.31 & 0.32 & 0.28 & 0.21 & 0.33 &  & 0.27 & 0.18 & 0.20 \\
    Jazz & 0.32 & 0.17 & 0.16 & 0.28 & 0.17 & 0.30 & 0.37 & 0.55 & 0.36 & 0.40 & 0.42 & 0.35 & 0.35 & 0.27 &  & 0.31 & 0.30 \\
    Latin & 0.25 & 0.19 & 0.10 & 0.21 & 0.24 & 0.37 & 0.36 & 0.31 & 0.23 & 0.27 & 0.28 & 0.17 & 0.28 & 0.18 & 0.31 &  & 0.27 \\
    Opera & 0.17 & 0.08 & 0.11 & 0.17 & 0.14 & 0.32 & 0.19 & 0.27 & 0.24 & 0.39 & 0.44 & 0.17 & 0.29 & 0.20 & 0.30 & 0.27 &  \\
    \bottomrule
    \end{tabular}
  \end{table}
  \end{landscape}

  \begin{landscape}
  \begin{table}
    \caption{Cramer's V between music genres (OpenAI)}
    \label{tab:cramersv_openai}
    \centering
    \footnotesize
    \begin{tabular}[t]{lrrrrrrrrrrrrrrrrr}
    \toprule
    & \rotatebox{90}{Pop music} & \rotatebox{90}{Hip-hop} & \rotatebox{90}{Country} & \rotatebox{90}{Alternative rock} & \rotatebox{90}{Electronic} & \rotatebox{90}{Ethnic music} & \rotatebox{90}{Reggae} & \rotatebox{90}{Blues} & \rotatebox{90}{Bluegrass} & \rotatebox{90}{Broadway} & \rotatebox{90}{Classical} & \rotatebox{90}{Classic rock} & \rotatebox{90}{Folk} & \rotatebox{90}{Gospel} & \rotatebox{90}{Jazz} & \rotatebox{90}{Latin} & \rotatebox{90}{Opera} \\
    \midrule
    Pop music &  & 0.40 & 0.34 & 0.37 & 0.35 & 0.24 & 0.21 & 0.04 & 0.39 & 0.19 & 0.08 & 0.12 & 0.29 & 0.29 & 0.11 & 0.21 & 0.10 \\
    Hip-hop & 0.40 &  & 0.60 & 0.26 & 0.68 & 0.35 & 0.55 & 0.20 & 0.64 & 0.08 & 0.25 & 0.31 & 0.70 & 0.38 & 0.19 & 0.29 & 0.11 \\
    Country & 0.34 & 0.60 &  & 0.23 & 0.59 & 0.38 & 0.41 & 0.26 & 0.85 & 0.13 & 0.03 & 0.29 & 0.58 & 0.55 & 0.32 & 0.24 & 0.14 \\
    Alternative rock & 0.37 & 0.26 & 0.23 &  & 0.40 & 0.05 & 0.08 & 0.00 & 0.19 & 0.07 & 0.02 & 0.05 & 0.03 & 0.52 & 0.02 & 0.03 & 0.09 \\
    Electronic & 0.35 & 0.68 & 0.59 & 0.40 &  & 0.35 & 0.30 & 0.15 & 0.54 & 0.04 & 0.09 & 0.27 & 0.43 & 0.56 & 0.15 & 0.24 & 0.07 \\
    Ethnic music & 0.24 & 0.35 & 0.38 & 0.05 & 0.35 &  & 0.44 & 0.11 & 0.41 & 0.07 & 0.10 & 0.29 & 0.35 & 0.19 & 0.21 & 0.19 & 0.10 \\
    Reggae & 0.21 & 0.55 & 0.41 & 0.08 & 0.30 & 0.44 &  & 0.17 & 0.42 & 0.06 & 0.17 & 0.30 & 0.54 & 0.06 & 0.21 & 0.18 & 0.06 \\
    Blues & 0.04 & 0.20 & 0.26 & 0.00 & 0.15 & 0.11 & 0.17 &  & 0.29 & 0.39 & 0.50 & 0.06 & 0.17 & 0.14 & 0.75 & 0.07 & 0.11 \\
    Bluegrass & 0.39 & 0.64 & 0.85 & 0.19 & 0.54 & 0.41 & 0.42 & 0.29 &  & 0.13 & 0.05 & 0.30 & 0.65 & 0.53 & 0.35 & 0.25 & 0.14 \\
    Broadway & 0.19 & 0.08 & 0.13 & 0.07 & 0.04 & 0.07 & 0.06 & 0.39 & 0.13 &  & 0.61 & 0.09 & 0.16 & 0.08 & 0.48 & 0.02 & 0.27 \\
    Classical & 0.08 & 0.25 & 0.03 & 0.02 & 0.09 & 0.10 & 0.17 & 0.50 & 0.05 & 0.61 &  & 0.12 & 0.31 & 0.05 & 0.64 & 0.14 & 0.22 \\
    Classic rock & 0.12 & 0.31 & 0.29 & 0.05 & 0.27 & 0.29 & 0.30 & 0.06 & 0.30 & 0.09 & 0.12 &  & 0.34 & 0.01 & 0.14 & 0.02 & 0.04 \\
    Folk & 0.29 & 0.70 & 0.58 & 0.03 & 0.43 & 0.35 & 0.54 & 0.17 & 0.65 & 0.16 & 0.31 & 0.34 &  & 0.27 & 0.18 & 0.30 & 0.08 \\
    Gospel & 0.29 & 0.38 & 0.55 & 0.52 & 0.56 & 0.19 & 0.06 & 0.14 & 0.53 & 0.08 & 0.05 & 0.01 & 0.27 &  & 0.14 & 0.15 & 0.01 \\
    Jazz & 0.11 & 0.19 & 0.32 & 0.02 & 0.15 & 0.21 & 0.21 & 0.75 & 0.35 & 0.48 & 0.64 & 0.14 & 0.18 & 0.14 &  & 0.02 & 0.19 \\
    Latin & 0.21 & 0.29 & 0.24 & 0.03 & 0.24 & 0.19 & 0.18 & 0.07 & 0.25 & 0.02 & 0.14 & 0.02 & 0.30 & 0.15 & 0.02 &  & 0.01 \\
    Opera & 0.10 & 0.11 & 0.14 & 0.09 & 0.07 & 0.10 & 0.06 & 0.11 & 0.14 & 0.27 & 0.22 & 0.04 & 0.08 & 0.01 & 0.19 & 0.01 &  \\
    \bottomrule
    \end{tabular}
  \end{table}
  \end{landscape}

  \begin{landscape}
  \begin{table}
    \caption{Cramer's V between music genres (Anthropic)}
    \label{tab:cramersv_anthropic}
    \centering
    \footnotesize
    \begin{tabular}[t]{lrrrrrrrrrrrrrrrrr}
    \toprule
    & \rotatebox{90}{Pop music} & \rotatebox{90}{Hip-hop} & \rotatebox{90}{Country} & \rotatebox{90}{Alternative rock} & \rotatebox{90}{Electronic} & \rotatebox{90}{Ethnic music} & \rotatebox{90}{Reggae} & \rotatebox{90}{Blues} & \rotatebox{90}{Bluegrass} & \rotatebox{90}{Broadway} & \rotatebox{90}{Classical} & \rotatebox{90}{Classic rock} & \rotatebox{90}{Folk} & \rotatebox{90}{Gospel} & \rotatebox{90}{Jazz} & \rotatebox{90}{Latin} & \rotatebox{90}{Opera} \\
    \midrule
    Pop music &  & 0.55 & 0.20 & 0.47 & 0.52 & 0.11 & 0.25 & 0.26 & 0.44 & 0.06 & 0.26 & 0.08 & 0.38 & 0.33 & 0.27 & 0.20 & 0.18 \\
    Hip-hop & 0.55 &  & 0.34 & 0.65 & 0.81 & 0.10 & 0.48 & 0.23 & 0.52 & 0.20 & 0.36 & 0.15 & 0.44 & 0.55 & 0.12 & 0.17 & 0.14 \\
    Country & 0.20 & 0.34 &  & 0.12 & 0.32 & 0.41 & 0.52 & 0.21 & 0.67 & 0.08 & 0.03 & 0.20 & 0.31 & 0.26 & 0.35 & 0.14 & 0.10 \\
    Alternative rock & 0.47 & 0.65 & 0.12 &  & 0.70 & 0.06 & 0.09 & 0.33 & 0.32 & 0.09 & 0.19 & 0.05 & 0.25 & 0.69 & 0.15 & 0.08 & 0.13 \\
    Electronic & 0.52 & 0.81 & 0.32 & 0.70 &  & 0.12 & 0.33 & 0.24 & 0.47 & 0.10 & 0.26 & 0.18 & 0.34 & 0.56 & 0.14 & 0.18 & 0.11 \\
    Ethnic music & 0.11 & 0.10 & 0.41 & 0.06 & 0.12 &  & 0.12 & 0.04 & 0.29 & 0.00 & 0.16 & 0.15 & 0.14 & 0.19 & 0.18 & 0.02 & 0.06 \\
    Reggae & 0.25 & 0.48 & 0.52 & 0.09 & 0.33 & 0.12 &  & 0.22 & 0.41 & 0.18 & 0.18 & 0.16 & 0.26 & 0.07 & 0.18 & 0.21 & 0.08 \\
    Blues & 0.26 & 0.23 & 0.21 & 0.33 & 0.24 & 0.04 & 0.22 &  & 0.04 & 0.13 & 0.45 & 0.00 & 0.40 & 0.24 & 0.59 & 0.13 & 0.21 \\
    Bluegrass & 0.44 & 0.52 & 0.67 & 0.32 & 0.47 & 0.29 & 0.41 & 0.04 &  & 0.14 & 0.26 & 0.14 & 0.53 & 0.44 & 0.10 & 0.34 & 0.01 \\
    Broadway & 0.06 & 0.20 & 0.08 & 0.09 & 0.10 & 0.00 & 0.18 & 0.13 & 0.14 &  & 0.56 & 0.06 & 0.48 & 0.19 & 0.02 & 0.11 & 0.30 \\
    Classical & 0.26 & 0.36 & 0.03 & 0.19 & 0.26 & 0.16 & 0.18 & 0.45 & 0.26 & 0.56 &  & 0.06 & 0.72 & 0.16 & 0.46 & 0.19 & 0.26 \\
    Classic rock & 0.08 & 0.15 & 0.20 & 0.05 & 0.18 & 0.15 & 0.16 & 0.00 & 0.14 & 0.06 & 0.06 &  & 0.09 & 0.01 & 0.06 & 0.01 & 0.02 \\
    Folk & 0.38 & 0.44 & 0.31 & 0.25 & 0.34 & 0.14 & 0.26 & 0.40 & 0.53 & 0.48 & 0.72 & 0.09 &  & 0.33 & 0.24 & 0.24 & 0.21 \\
    Gospel & 0.33 & 0.55 & 0.26 & 0.69 & 0.56 & 0.19 & 0.07 & 0.24 & 0.44 & 0.19 & 0.16 & 0.01 & 0.33 &  & 0.06 & 0.16 & 0.06 \\
    Jazz & 0.27 & 0.12 & 0.35 & 0.15 & 0.14 & 0.18 & 0.18 & 0.59 & 0.10 & 0.02 & 0.46 & 0.06 & 0.24 & 0.06 &  & 0.09 & 0.36 \\
    Latin & 0.20 & 0.17 & 0.14 & 0.08 & 0.18 & 0.02 & 0.21 & 0.13 & 0.34 & 0.11 & 0.19 & 0.01 & 0.24 & 0.16 & 0.09 &  & 0.05 \\
    Opera & 0.18 & 0.14 & 0.10 & 0.13 & 0.11 & 0.06 & 0.08 & 0.21 & 0.01 & 0.30 & 0.26 & 0.02 & 0.21 & 0.06 & 0.36 & 0.05 &  \\
    \bottomrule
    \end{tabular}
  \end{table}
  \end{landscape}

  \begin{landscape}
  \begin{table}
    \caption{Cramer's V between music genres (DeepSeek)}
    \label{tab:cramersv_deepseek}
    \centering
    \footnotesize
    \begin{tabular}[t]{lrrrrrrrrrrrrrrrrr}
    \toprule
    & \rotatebox{90}{Pop music} & \rotatebox{90}{Hip-hop} & \rotatebox{90}{Country} & \rotatebox{90}{Alternative rock} & \rotatebox{90}{Electronic} & \rotatebox{90}{Ethnic music} & \rotatebox{90}{Reggae} & \rotatebox{90}{Blues} & \rotatebox{90}{Bluegrass} & \rotatebox{90}{Broadway} & \rotatebox{90}{Classical} & \rotatebox{90}{Classic rock} & \rotatebox{90}{Folk} & \rotatebox{90}{Gospel} & \rotatebox{90}{Jazz} & \rotatebox{90}{Latin} & \rotatebox{90}{Opera} \\
    \midrule
    Pop music &  & 0.45 & 0.32 & 0.40 & 0.44 & 0.09 & 0.31 & 0.22 & 0.40 & 0.05 & 0.39 & 0.18 & 0.39 & 0.30 & 0.35 & 0.23 & 0.30 \\
    Hip-hop & 0.45 &  & 0.64 & 0.23 & 0.73 & 0.14 & 0.68 & 0.11 & 0.64 & 0.17 & 0.37 & 0.37 & 0.53 & 0.48 & 0.02 & 0.24 & 0.16 \\
    Country & 0.32 & 0.64 &  & 0.19 & 0.56 & 0.25 & 0.60 & 0.02 & 0.80 & 0.03 & 0.14 & 0.37 & 0.39 & 0.49 & 0.22 & 0.16 & 0.06 \\
    Alternative rock & 0.40 & 0.23 & 0.19 &  & 0.42 & 0.07 & 0.02 & 0.25 & 0.16 & 0.02 & 0.10 & 0.05 & 0.11 & 0.52 & 0.17 & 0.05 & 0.13 \\
    Electronic & 0.44 & 0.73 & 0.56 & 0.42 &  & 0.14 & 0.46 & 0.15 & 0.52 & 0.06 & 0.27 & 0.38 & 0.39 & 0.60 & 0.06 & 0.21 & 0.13 \\
    Ethnic music & 0.09 & 0.14 & 0.25 & 0.07 & 0.14 &  & 0.03 & 0.13 & 0.30 & 0.00 & 0.07 & 0.26 & 0.16 & 0.21 & 0.04 & 0.05 & 0.03 \\
    Reggae & 0.31 & 0.68 & 0.60 & 0.02 & 0.46 & 0.03 &  & 0.12 & 0.53 & 0.09 & 0.25 & 0.33 & 0.37 & 0.19 & 0.22 & 0.21 & 0.11 \\
    Blues & 0.22 & 0.11 & 0.02 & 0.25 & 0.15 & 0.13 & 0.12 &  & 0.03 & 0.27 & 0.33 & 0.02 & 0.37 & 0.20 & 0.36 & 0.01 & 0.11 \\
    Bluegrass & 0.40 & 0.64 & 0.80 & 0.16 & 0.52 & 0.30 & 0.53 & 0.03 &  & 0.06 & 0.26 & 0.31 & 0.53 & 0.47 & 0.09 & 0.43 & 0.01 \\
    Broadway & 0.05 & 0.17 & 0.03 & 0.02 & 0.06 & 0.00 & 0.09 & 0.27 & 0.06 &  & 0.52 & 0.10 & 0.46 & 0.05 & 0.27 & 0.11 & 0.33 \\
    Classical & 0.39 & 0.37 & 0.14 & 0.10 & 0.27 & 0.07 & 0.25 & 0.33 & 0.26 & 0.52 &  & 0.15 & 0.71 & 0.11 & 0.63 & 0.25 & 0.32 \\
    Classic rock & 0.18 & 0.37 & 0.37 & 0.05 & 0.38 & 0.26 & 0.33 & 0.02 & 0.31 & 0.10 & 0.15 &  & 0.26 & 0.04 & 0.11 & 0.01 & 0.02 \\
    Folk & 0.39 & 0.53 & 0.39 & 0.11 & 0.39 & 0.16 & 0.37 & 0.37 & 0.53 & 0.46 & 0.71 & 0.26 &  & 0.27 & 0.35 & 0.33 & 0.24 \\
    Gospel & 0.30 & 0.48 & 0.49 & 0.52 & 0.60 & 0.21 & 0.19 & 0.20 & 0.47 & 0.05 & 0.11 & 0.04 & 0.27 &  & 0.09 & 0.17 & 0.10 \\
    Jazz & 0.35 & 0.02 & 0.22 & 0.17 & 0.06 & 0.04 & 0.22 & 0.36 & 0.09 & 0.27 & 0.63 & 0.11 & 0.35 & 0.09 &  & 0.21 & 0.31 \\
    Latin & 0.23 & 0.24 & 0.16 & 0.05 & 0.21 & 0.05 & 0.21 & 0.01 & 0.43 & 0.11 & 0.25 & 0.01 & 0.33 & 0.17 & 0.21 &  & 0.08 \\
    Opera & 0.30 & 0.16 & 0.06 & 0.13 & 0.13 & 0.03 & 0.11 & 0.11 & 0.01 & 0.33 & 0.32 & 0.02 & 0.24 & 0.10 & 0.31 & 0.08 &  \\
    \bottomrule
    \end{tabular}
  \end{table}
  \end{landscape}

  \begin{landscape}
  \begin{table}
    \caption{Cramer's V between music genres (Bootstrap)}
    \label{tab:cramersv_bootstrap}
    \centering
    \footnotesize
    \begin{tabular}[t]{lrrrrrrrrrrrrrrrrr}
    \toprule
    & \rotatebox{90}{Pop music} & \rotatebox{90}{Hip-hop} & \rotatebox{90}{Country} & \rotatebox{90}{Alternative rock} & \rotatebox{90}{Electronic} & \rotatebox{90}{Ethnic music} & \rotatebox{90}{Reggae} & \rotatebox{90}{Blues} & \rotatebox{90}{Bluegrass} & \rotatebox{90}{Broadway} & \rotatebox{90}{Classical} & \rotatebox{90}{Classic rock} & \rotatebox{90}{Folk} & \rotatebox{90}{Gospel} & \rotatebox{90}{Jazz} & \rotatebox{90}{Latin} & \rotatebox{90}{Opera} \\
    \midrule
    Pop music &  & 0.33 & 0.28 & 0.37 & 0.27 & 0.21 & 0.33 & 0.36 & 0.25 & 0.32 & 0.26 & 0.42 & 0.27 & 0.17 & 0.32 & 0.25 & 0.17 \\
    Hip-hop & 0.33 &  & 0.09 & 0.32 & 0.34 & 0.16 & 0.37 & 0.27 & 0.07 & 0.08 & 0.06 & 0.16 & 0.07 & 0.07 & 0.18 & 0.19 & 0.08 \\
    Country & 0.28 & 0.09 &  & 0.12 & 0.06 & 0.09 & 0.13 & 0.20 & 0.33 & 0.22 & 0.14 & 0.33 & 0.26 & 0.24 & 0.15 & 0.10 & 0.11 \\
    Alternative rock & 0.37 & 0.32 & 0.12 &  & 0.31 & 0.25 & 0.38 & 0.32 & 0.26 & 0.20 & 0.23 & 0.34 & 0.27 & 0.07 & 0.28 & 0.21 & 0.17 \\
    Electronic & 0.27 & 0.34 & 0.06 & 0.31 &  & 0.22 & 0.30 & 0.20 & 0.13 & 0.17 & 0.15 & 0.15 & 0.15 & 0.06 & 0.17 & 0.24 & 0.14 \\
    Ethnic music & 0.21 & 0.16 & 0.09 & 0.25 & 0.22 &  & 0.32 & 0.31 & 0.26 & 0.27 & 0.31 & 0.17 & 0.32 & 0.19 & 0.30 & 0.37 & 0.32 \\
    Reggae & 0.33 & 0.37 & 0.13 & 0.38 & 0.30 & 0.32 &  & 0.44 & 0.28 & 0.24 & 0.24 & 0.28 & 0.28 & 0.17 & 0.37 & 0.36 & 0.20 \\
    Blues & 0.36 & 0.27 & 0.20 & 0.32 & 0.20 & 0.31 & 0.44 &  & 0.44 & 0.36 & 0.35 & 0.40 & 0.42 & 0.32 & 0.55 & 0.31 & 0.27 \\
    Bluegrass & 0.25 & 0.07 & 0.33 & 0.26 & 0.13 & 0.26 & 0.28 & 0.44 &  & 0.35 & 0.32 & 0.34 & 0.60 & 0.31 & 0.36 & 0.23 & 0.24 \\
    Broadway & 0.32 & 0.08 & 0.22 & 0.20 & 0.17 & 0.27 & 0.24 & 0.36 & 0.35 &  & 0.47 & 0.34 & 0.40 & 0.32 & 0.40 & 0.28 & 0.39 \\
    Classical & 0.26 & 0.06 & 0.14 & 0.23 & 0.15 & 0.31 & 0.24 & 0.35 & 0.32 & 0.47 &  & 0.30 & 0.39 & 0.28 & 0.42 & 0.28 & 0.44 \\
    Classic rock & 0.42 & 0.16 & 0.33 & 0.34 & 0.15 & 0.17 & 0.28 & 0.40 & 0.34 & 0.34 & 0.30 &  & 0.33 & 0.21 & 0.35 & 0.17 & 0.17 \\
    Folk & 0.27 & 0.07 & 0.26 & 0.27 & 0.15 & 0.32 & 0.28 & 0.42 & 0.60 & 0.40 & 0.39 & 0.33 &  & 0.33 & 0.35 & 0.28 & 0.29 \\
    Gospel & 0.17 & 0.07 & 0.24 & 0.07 & 0.06 & 0.19 & 0.17 & 0.32 & 0.31 & 0.32 & 0.28 & 0.21 & 0.33 &  & 0.27 & 0.18 & 0.20 \\
    Jazz & 0.32 & 0.18 & 0.15 & 0.28 & 0.17 & 0.30 & 0.37 & 0.55 & 0.36 & 0.40 & 0.42 & 0.35 & 0.35 & 0.27 &  & 0.31 & 0.31 \\
    Latin & 0.25 & 0.19 & 0.10 & 0.21 & 0.24 & 0.37 & 0.36 & 0.31 & 0.23 & 0.28 & 0.28 & 0.17 & 0.28 & 0.18 & 0.31 &  & 0.27 \\
    Opera & 0.17 & 0.08 & 0.11 & 0.17 & 0.14 & 0.32 & 0.20 & 0.27 & 0.24 & 0.39 & 0.44 & 0.17 & 0.29 & 0.20 & 0.31 & 0.27 &  \\
    \bottomrule
    \end{tabular}
  \end{table}
  \end{landscape}

\end{document}